\RequirePackage{xcolor}
\documentclass{article}

\usepackage{microtype}
\usepackage{booktabs} 
\usepackage{tabularx}

\usepackage{hyperref}

\usepackage[accepted]{icml2019}

\usepackage{subfig}
\usepackage{caption}
\usepackage{amssymb}
\usepackage{amsmath}
\usepackage[title]{appendix}
\usepackage{moresize}
\usepackage{setspace}
\usepackage{algpseudocode}
\usepackage[normalem]{ulem}
\usepackage{graphicx}
\graphicspath{{figs/}}

\usepackage[subpreambles=true]{standalone}
\usepackage{tikz}
\usetikzlibrary{bayesnet}
\usetikzlibrary{fit}
\usepackage{anyfontsize}
\usetikzlibrary{positioning}
\usetikzlibrary{shapes,snakes}
\usetikzlibrary{arrows}
\usetikzlibrary{shapes.misc}
\usetikzlibrary{calc}
\usetikzlibrary{arrows.meta}

\usepackage{amsmath,amsfonts,bm}

\def\eqref#1{equation~\ref{#1}}

\def\1{\bm{1}}

\def\vtheta{{\bm{\theta}}}

\def\vx{{\bm{x}}}
\def\vy{{\bm{y}}}

\DeclareMathAlphabet{\mathsfit}{\encodingdefault}{\sfdefault}{m}{sl}
\SetMathAlphabet{\mathsfit}{bold}{\encodingdefault}{\sfdefault}{bx}{n}

\newcommand{\E}{\mathbb{E}}

\newcommand{\KL}{D_{\mathrm{KL}}}

\newcommand{\dint}{\text{d}}
\newcommand{\vphi}{\boldsymbol{\phi}}
\newcommand{\vpi}{\boldsymbol{\pi}}
\newcommand{\vpsi}{\boldsymbol{\psi}}
\newcommand{\vomg}{\boldsymbol{\omega}}

\renewcommand{\vx}{\mathbf{x}}
\renewcommand{\vy}{\mathbf{y}}

\renewcommand{\b}{\mathbf}

\newcommand{\mvn}{\mathcal{MN}}

\newcommand{\diag}[1]{\text{diag}(#1)}

\renewcommand{\E}{\mathbb{E}}
\newcommand{\D}{\mathcal{D}}

\usepackage{cleveref}

\icmltitlerunning{Predictive Uncertainty Quantification with Compound Density Networks}

\begin{document}

\twocolumn[
\icmltitle{Predictive Uncertainty Quantification with Compound Density Networks}



\icmlsetsymbol{equal}{*}

\begin{icmlauthorlist}
\icmlauthor{Agustinus Kristiadi}{tue}
\icmlauthor{Sina D\"{a}ubener}{bochum}
\icmlauthor{Asja Fischer}{bochum}
\end{icmlauthorlist}

\icmlaffiliation{tue}{Department of Computer Science, University of T\"{u}bingen, Germany}
\icmlaffiliation{bochum}{Department of Mathematics, Ruhr University Bochum, Germany}

\icmlcorrespondingauthor{Agustinus Kristiadi}{agustinus.kristiadi@uni-tuebingen.de}

\icmlkeywords{Machine Learning, ICML}

\vskip 0.3in
]



\printAffiliationsAndNotice{}  

\begin{abstract}
Despite the huge success of deep neural networks (NNs), finding good mechanisms for quantifying their prediction uncertainty is still an open problem. Bayesian neural networks are one of the most popular approaches to uncertainty quantification. On the other hand, it was recently shown that ensembles of NNs, which belong to the class of mixture models, can be used to quantify prediction uncertainty. In this paper, we build upon these two approaches. First, we increase the mixture model's flexibility by replacing the fixed mixing weights by an adaptive, input-dependent distribution (specifying the probability of each component) represented by NNs, and by considering uncountably many mixture components. The resulting class of models can be seen as the continuous counterpart to mixture density networks and is therefore referred to as \emph{compound density networks} (CDNs). We employ both maximum likelihood and variational Bayesian inference to train CDNs, and empirically show that they yield better uncertainty estimates on out-of-distribution data and are more robust to adversarial examples than the previous approaches.
\end{abstract}

\section{Introduction}
\label{sec:intro}

Deep neural networks (NNs) have achieved state-of-the-art performance in many application areas, such as computer vision \citep{krizhevsky2012imagenet} and natural language processing \citep{collobert2011natural}. However, despite achieving impressive prediction accuracy on these supervised machine learning tasks, NNs do not provide good ways of quantifying predictive uncertainty. This is undesirable for many mission-critical applications, where taking wrong predictions with high confidence could have fatal consequences (e.g.~in medical diagnostics or autonomous driving). 

A principled and the most explored way to quantify the uncertainty in NNs is through Bayesian inference. In the so-called Bayesian neural networks (BNNs) \citep{neal1995bayesian}, the NN parameters are treated as random variables and the goal of learning is to infer the posterior probability distribution of the parameters given the training data. Since exact Bayesian inference in NNs is computationally intractable, different approximation techniques have been proposed \citep[etc.]{neal1995bayesian, blundell_weight_2015, hernandez2015probabilistic, ritter_scalable_2018}. Given the (approximate) posterior, the final predictive distribution is obtained as the expected predictive distribution under the posterior. This expectation can be seen as an ensemble of an uncountably infinite number of predictors, where the prediction of each model is weighted by the posterior probability of the corresponding parameters.

Based on a Bayesian interpretation of dropout \citep{srivastava2014dropout}, \citet{gal2016dropout} proposed to apply it not only during training but also when making predictions to estimate predictive uncertainty. Interestingly, dropout has been also interpreted as ensemble model \citep{srivastava2014dropout} where the predictions are averaged over the different NNs resulting from different dropout-masks. Inspired by this, \citet{lakshminarayanan2017simple} proposed to use a simple NN ensemble to quantify the prediction uncertainty, i.e.~to train a set of independent NNs using a proper scoring rule and defining the final prediction as the arithmetic mean of the outputs of the individual models, which corresponds to defining a uniformly-weighted mixture model. It is argued, that the model is able to encode two sources of uncertainty by calibrating the target uncertainty (i.e.~uncertainty in target $\vy$ given input $\vx$) in each component and capturing the model uncertainty by averaging over the components. 

In this paper, we therefore aim at further investigating the potential that lies in employing mixture distributions for uncertainty quantification. 
The flexibility of the mixture model can be increased by learning input-conditioned mixture weights like it is done by mixture density networks (MDNs) \citep{bishop1994mixture}. Furthermore, one can consider uncountably many component distributions instead of a finite set, which turns the mixture distribution into a compound distribution. We combine both by deriving the continuous counterpart of MDNs, which we call \emph{compound density networks} (CDNs). 
These networks can be trained by likelihood maximization. Moreover,
variational Bayes 
can be employed to infer the posterior distribution over the CDN parameters,
leading to a combination of the mixture model and the Bayesian approach to uncertainty modeling. 
We experimentally show that CDNs allow for 
better uncertainty quantification and are more robust to adversarial examples than previous approaches.

This paper is organized as follows. In \Cref{sec:mdn} we give a brief introduction to MDNs. We then formally define CDNs in \Cref{sec:cdn}. We review related work in \Cref{sec:related} and present a detailed experimental analysis in \Cref{sec:experiment}. Finally, we conclude our paper in \Cref{sec:conclusion}.

\section{Mixture Density Networks}
\label{sec:mdn}

Let $\D = \{ \vx_n, \vy_n \}_{n=1}^N$ be an i.i.d~dataset. Let us define the following conditional mixture model
\begin{equation}
    p(\vy \vert \vx) = \sum_{k=1}^K  p\left(\vy; \vphi_k(\vx) \right) p(\vphi_k(\vx); \vpi(\vx)) \, , \label{eq:mdn_model}
\end{equation}
%
and an NN that maps $\vx$ onto both the parameters $\vpi(\vx)$  of the  mixing distribution and the parameters $ \{ \vphi_k (\vx) \}_{k=1}^K$ of the $K$ mixture components.
The complete system is called mixture density network (MDN) and was proposed by \citet{bishop1994mixture}. That is, an MDN is an NN parametrizing a conditional mixture distribution, where both the mixture components  and the mixture coefficients depend on input $\vx$.\footnote{For instance, as in the original publication, the mixture components could be $K$ Gaussians, with $\vphi_k(\vx)$ being input-specific means and variances, and the mixture probabilities could be given by (applying the softmax function to the unnormalized) mixing weights $\vpi(\vx)$, both computed by one NN.} MDNs can be trained by maximizing the log-likelihood of the parameters of the NN given the training set $\mathcal{D}$ using gradient-based optimizers such as stochastic gradient descent (SGD) and its variants.

MDNs belong to a broader class of models called mixture of experts (MoE) \citep{jacobs1991adaptive} which differ from standard mixture models by assuming that the mixture coefficients depend on the input.\footnote{See \citet[ch.~5.6 and ch.~14.5.3]{Bishop:2006} and \citet[ch.~11.2.4]{Murphy:2012:MLP:2380985} for a detailed discussion of MDNs.} Because of its formulation as a mixture distribution, the predictive distribution of an MDN can handle multimodality better than a standard discriminative neural network.

\section{Compound density networks}
\label{sec:cdn}

We aim at generalizing the MDN from a finite mixture  distribution to a mixture of an uncountable set of components. The continuous counterpart of a  conditional mixture distribution in \cref{eq:mdn_model} is given by the conditional \emph{compound probability distribution}
\begin{align}
    p(\vy \vert \vx)
        = \int p(\vy; \vphi(\vx)) p(\vphi(\vx); \vpi(\vx)) \, \dint{\vphi(\vx)} \label{eq:cpd_mixture} \, ,
\end{align}
where $\vphi(\vx)$ turns from a discrete into a continuous random variable.


We now want to follow the approach of MDNs to model the parameters of the components and the mixing distribution by NNs. To handle the continuous state space of $\vphi(\vx)$ in the case of a compound distribution, the key idea is now to let $\vphi(\vx)$ be given by a stochastic NN $f(\vx; \vtheta)=\vphi(\vx)$ with stochastic parameters $\vtheta$. Since given $\vx$, $f$ is a deterministic map from $\vtheta$ to $\vphi(\vx)$, it is possible to replace the mixing distribution $p(\vphi(\vx); \vpi(\vx))=p(f(\vx;\vtheta);  \vpi(\vx))$ by a distribution $p(\vtheta; \vpi(\vx))$ over $\vtheta$. We further assume, that the parameters $\vpi(\vx)$ of the mixing distribution are given by some parametrized function $g(\vx; \vpsi) = \vpi(\vx)$ which can also be modeled based on NNs. In correspondence to MDNs, the complete system is called \emph{compound density network} (CDN) and it is summarized by the following equation
\begin{align}
    p(\vy \vert \vx; \vpsi)
        &:= \int p(\vy; f(\vx; \vtheta)) p(\vtheta; g(\vx; \vpsi)
        ) \, \dint{\vtheta} \nonumber \\
        &= \mathbb{E}_{p(\vtheta; g(\vx; \vpsi))} [\,p(\vy; f(\vx; \vtheta))] \enspace . \label{eq:cdn_mixture} 
\end{align}
%

As MDNs, CDNs can be trained by maximizing the log-likelihood  
%
of $\vpsi$ given the data set $\mathcal{D}$.
%
%
Moreover, one can add a regularization term encouraging the mixing distribution to stay close to some distribution $p(\vtheta)$, which leads to the objective
\begin{align}
   \mathcal{L}_{\text{ML}}(\vpsi) :=
    &\sum_{n=1}^N \log \mathbb{E}_{p(\vtheta; g(\vx_n; \vpsi))} [\,p(\vy_n; f(\vx_n; \vtheta)] 
    \nonumber \\ 
            &- \lambda \sum_{n=1}^N\KL[p(\vtheta; g(\vx_n; \vpsi)) \Vert p(\vtheta)] \enspace ,
            \label{eq:MLobjective}
\end{align}
where $\lambda$ is a hyperparameter controlling the strength of the regularization.\footnote{Note, that the objective gets equivalent to the one proposed by \citet{alemi2017deep} when it is approximated based on a single sample of $\vtheta$.
We experimentally show that this objective leads to better results than theirs (when approximated with more samples) in the supplement.}

\begin{algorithm}[htb]
  \caption{The training procedure of CDNs with $\mathcal{L}_\text{ML}$.}
  \label{algo:cdn_learning}
  \begin{algorithmic}[1]
        \Require
            \Statex Mini-batch size $M$, number of samples $S$ of $\vtheta$, regularization strength $\lambda$, and learning rate $\alpha$.
        \setstretch{1.05}
        \While{the stopping criterion is not satisfied}
            \State $\{ \vx_m, \vy_m \}_{m=1}^M \sim \mathcal{D}$ 
            \For{$m = 1, \dots, M$; $s = 1, \dots, S$}
                \State $\vtheta_{ms} \sim p(\vtheta; g(\vx_m; \vpsi))$ 
                \State $\vphi_s(\vx_m) = f(\vx_m; \vtheta_{ms})$ 
            \EndFor
            \State $\mathcal{L}(\vpsi) = \sum_{m=1}^M \log \frac{1}{S} \sum_{s=1}^S p(\vy_m; \vphi_s(\vx_m)) - \lambda \sum_{m=1}^M \KL[\, p(\vtheta; g(\vx_m; \vpsi)) \Vert p(\vtheta))]$
            \State $\vpsi \leftarrow \vpsi + \alpha \nabla \mathcal{L}(\vpsi)$ 
        \EndWhile
  \end{algorithmic}
\end{algorithm}

Alternatively, we can turn CDNs into Bayesian models by defining a prior $p(\vpsi)$ over $\vpsi$ 
and employing variational Bayes (VB) 
\citep{hinton1993keeping}  to  infer an approximate posterior $q(\vpsi; \vomg) \approx p(\vpsi \vert \D)$ by maximizing the  evidence lower bound (ELBO):
%
%
%
\begin{align}
  \mathcal{L}_{\text{VB}}(\vomg) := &
  \sum_{n=1}^N   \E_{q(\vpsi; \vomg)}[\log \mathbb{E}_{p(\vtheta; g(\vx_n; \vpsi))} [\,p(\vy_n; f(\vx_n; \vtheta)]] \nonumber  \\
        &- \KL[ q(\vpsi ; \vomg) \Vert\, p(\vpsi)] \enspace . \label{eq:CDN_ELBO}
\end{align}
%
Given the approximate posterior the predictive distribution of the Bayesian CDN is defined as
\begin{align}
  p(\vy \vert \vx) &= \int p(\vy \vert \vx; \vpsi) q(\vpsi; \vomg) \, \dint{\vpsi} \, , 
  \label{eq:bayesian_cdn}
\end{align}
which is approximated based on samples of $\vpsi$ and $\vtheta$.

We present pseudocode for training CDNs with $\mathcal{L}_\text{ML}$ in \Cref{algo:cdn_learning}.\footnote{Pseudocode for training CDNs with $\mathcal{L}_\text{VB}$ is presented in the supplement.} Note that CDNs correspond to an abstract framework for modeling compound distributions with NNs, i.e.~we still need to concretely define the stochastic NN $f(\vx; \vtheta)$ and choose the statistical models for the mixture components and the mixing distribution. 

\subsection{Probabilistic hypernetworks}
\label{subsec:prob_hypernet}

\citet{ha_hypernetworks_2016} proposed to (deterministically) generate the parameters of an NN by another NN, which they call the \emph{hypernetwork}\footnote{Specifically, they propose to apply a hypernetwork to compute the weight matrix of a recurrent NN at each time-step, given the current input and the previous hidden state.}. We would like to follow this approach for modeling a CDN, that is, we aim at modeling the mixing distribution  $p(\vtheta; g(\vx;\vpsi))$ over network parameters by NNs. Since now the hypernetworks map $\vx$ to a distribution over parameters instead of a specific value $\vtheta$, we refer to them as \emph{probabilistic hypernetworks}. In the following, we will describe this idea in more detail.

\begin{figure}
    \centering
    
    \begin{tikzpicture}[scale=3, node distance=8cm, auto, align=center, every node/.style = {draw=blue!60, fill=cyan!10, minimum height=2cm, minimum width=3cm}]

    \Huge
    
    \node(h) [draw, very thick, minimum width=1.5cm, minimum height=4cm, xshift=1cm] {$\b{h}$};
    \node(x) [draw, very thick, left=of h, minimum width=1.5cm, minimum height=6cm] {$\vx$};
    \node(y) [draw, very thick, right=of h, minimum width=1.5cm, minimum height=2cm, xshift=1cm] {$\vphi(\vx)$};
    
    \draw[very thick, -{Latex[length=3mm,width=5mm]}] (x) -- (h) node(w1)[midway, below, fill=none, draw=none] {$\b{W}_1 \sim p(\b{W}_1; \vpi_1(\vx))$};    
    \draw[very thick, -{Latex[length=3mm,width=5mm]}] (h) -- (y) node(w2)[midway, below, fill=none, draw=none] {$\b{W}_2 \sim p(\b{W}_2; \vpi_2(\b{h}))$};
    
    \node(g1) [draw=red!60, fill=red!5, very thick, below=of w1, yshift=4cm] {$\vpi_1(\vx) = g_1(\vx; \vpsi_1)$};
    \node(g2) [draw=red!60, fill=red!5, very thick, below=of w2, yshift=4cm] {$\vpi_2(\b{h}) = g_2(\b{h}; \vpsi_2)$};
    
    \draw[very thick, -{Latex[length=3mm,width=5mm]}] (g1) -- (w1);    
    \draw[very thick, -{Latex[length=3mm,width=5mm]}] (g2) -- (w2);
    
    \draw[very thick, bend right=55, -{Latex[length=3mm,width=5mm]}] (x) to (g1);
    \draw[very thick, bend right=55, -{Latex[length=3mm,width=5mm]}] (h) to (g2);

\end{tikzpicture}
    
    \caption{An example of a probabilistic hypernetwork applied to a two-layer MLP.}
    \label{fig:prob_hypernets}
\end{figure}
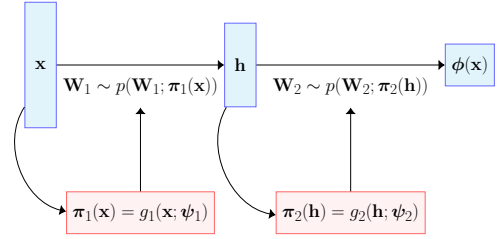

Let $f$ be a multi-layer perceptron
(MLP) with $L$-layers, parametrized by a set of layers' weight matrices\footnote{We assume that the bias parameters are absorbed into the weight matrix.} $\vtheta = \{ \b{W}_l \}_{l=1}^L$, that computes the parameters $\vphi(\vx) = f(\vx; \vtheta)$ of the CDNs component distribution in \cref{eq:cdn_mixture}. Let $\b{h}_1, \dots, \b{h}_{l-1}$ denote the states of the hidden layers, and let us define $\b{h}_0 = \vx$, and $\b{h}_L = f(\vx; \vtheta)$. We now assume the weight matrices  $\{\b{W}_l\}_{l=1}^L$ to be random variables and to be independent of each other given the state of the previous hidden layer. We define a series of probabilistic hypernetworks $g=\{g_l\}_{l=1}^L$ (parametrized by $\vpsi = \{\vpsi_l\}_{l=1}^L$), where $g_l$ maps $\b{h}_{l-1}$ to the parameters of the distribution of $\b{W}_l$, and let the joint distribution over $\vtheta$ be given by
\begin{equation}
    p(\vtheta; g(\vx; \vpsi)) := \prod_{l=1}^L p(\b{W}_l; g_l(\b{h}_{l-1}; \vpsi_l)) \label{eq:prob_hypernet_mixing} \, .
\end{equation}
An illustration of a stochastic two-layer network $f(\vx; \vtheta)$ computing $\vphi(\vx)$, where the distribution of the parameters is given by probabilistic hypernetworks as defined in \cref{eq:prob_hypernet_mixing}, is given in \Cref{fig:prob_hypernets}.



\subsection{Probabilistic hypernetworks with matrix variate normals}
\label{subsec:prob_hypernets_mvn}

A statistical model that was recently applied as the posterior over weight matrices in BNNs \citep{louizos_structured_2016, sun_learning_2017, DBLP:journals/corr/abs-1712-02390, ritter_scalable_2018} is the matrix variate normal (MVN) distribution \citep{gupta1999matrix}. An MVN is parametrized by three matrices: a mean matrix $\b{M}$ and two covariance factor matrices $\b{A}$ and $\b{B}$. It is connected to the multivariate Gaussian by the following equivalence
\begin{align}
    \b{X} &\sim \mvn(\b{X}; \b{M}, \b{A}, \b{B}) \nonumber \\
    &\iff \text{vec}(\b{X}) \sim \mathcal{N}(\text{vec}(\b{X}); \text{vec}(\b{M}), \b{B} \otimes \b{A}) \, , \label{eq:mvn}
\end{align}
where $\text{vec}(\b{X})$ denotes the vectorization of matrix $\b{X}$. Due to the Kronecker factorization of the covariance, an MVN requires fewer parameters compared to a multivariate Gaussian, which motivates us to use it as the distribution over weight matrices in this work. Furthermore, we assume that the covariance factor matrices are diagonal matrices, following \citet{louizos_structured_2016}. That is, we choose the mixture distribution of the CDN to be
\begin{align}
\label{eq:MVN}
    p(\vtheta; g(\vx; \vpsi)) &= \prod_{l=1}^L \mvn(\b{W}_l; g_l(\b{h}_{l-1}; \vpsi_l)) \\
        &= \prod_{l=1}^L \mvn(\b{W}_l; \b{M}_l, \diag{\b{a}_l}, \diag{\b{b}_l}) \nonumber \enspace,
\end{align}
where $g_l$ maps the state $\b{h}_{l-1}$ of the previous hidden layer onto the $l$-th MVN's parameters $ \{\b{M}_l, \b{a}_l, \b{b}_l\}$ defining the distribution over $\b{W}_l$. Suppose $\b{W}_l \in \mathbb{R}^{r \times c}$, then the corresponding MVN distribution has $rc + r + c$ parameters, which is more efficient compared to $rc + rc$ parameters when modeling $\b{W}_l$ as fully-factorized Gaussian random variable. 
To further reduce the model complexity  we use a vector-scaling parametrization similar to the one used by \citet{ha_hypernetworks_2016} and \citet{krueger_bayesian_2017} for the mean matrices $\{ \b{M}_l \}_{l=1}^L$. 
We detail this parametrization in the supplement.

For the regularization during ML training (\cref{eq:MLobjective}), we define the prior over $\vtheta$ to be $p(\vtheta) := \prod_{l=1}^L \mvn(\b{W}_l; \b{0}, \b{I}, \b{I})$.
Meanwhile, to perform variational Bayes (\cref{eq:CDN_ELBO}),
the prior over $\vpsi$ is assumed to be the product of standard MVNs (similar to $p(\vtheta)$), while the variational posterior is defined as product of diagonal MVNs with variational parameters $\vomg$, following \citet{louizos_structured_2016}. That is, for CDNs trained with VB, we assume  each 
probabilistic hypernetwork $g_l$ to be a variational matrix Gaussian (VMG), the BNN proposed by \citet{louizos_structured_2016}.


Note that using the mixing distribution and approximate posterior as defined above allows us to apply the reparametrization trick \citep{louizos_structured_2016}. Furthermore, the KL-divergence term in \cref{eq:CDN_ELBO} can be computed in closed form, as also noted by \citet{louizos_structured_2016}.

\begin{figure*}[t!]
    \centering

    \subfloat[ML-CDN]{\includegraphics[width=0.16\linewidth]{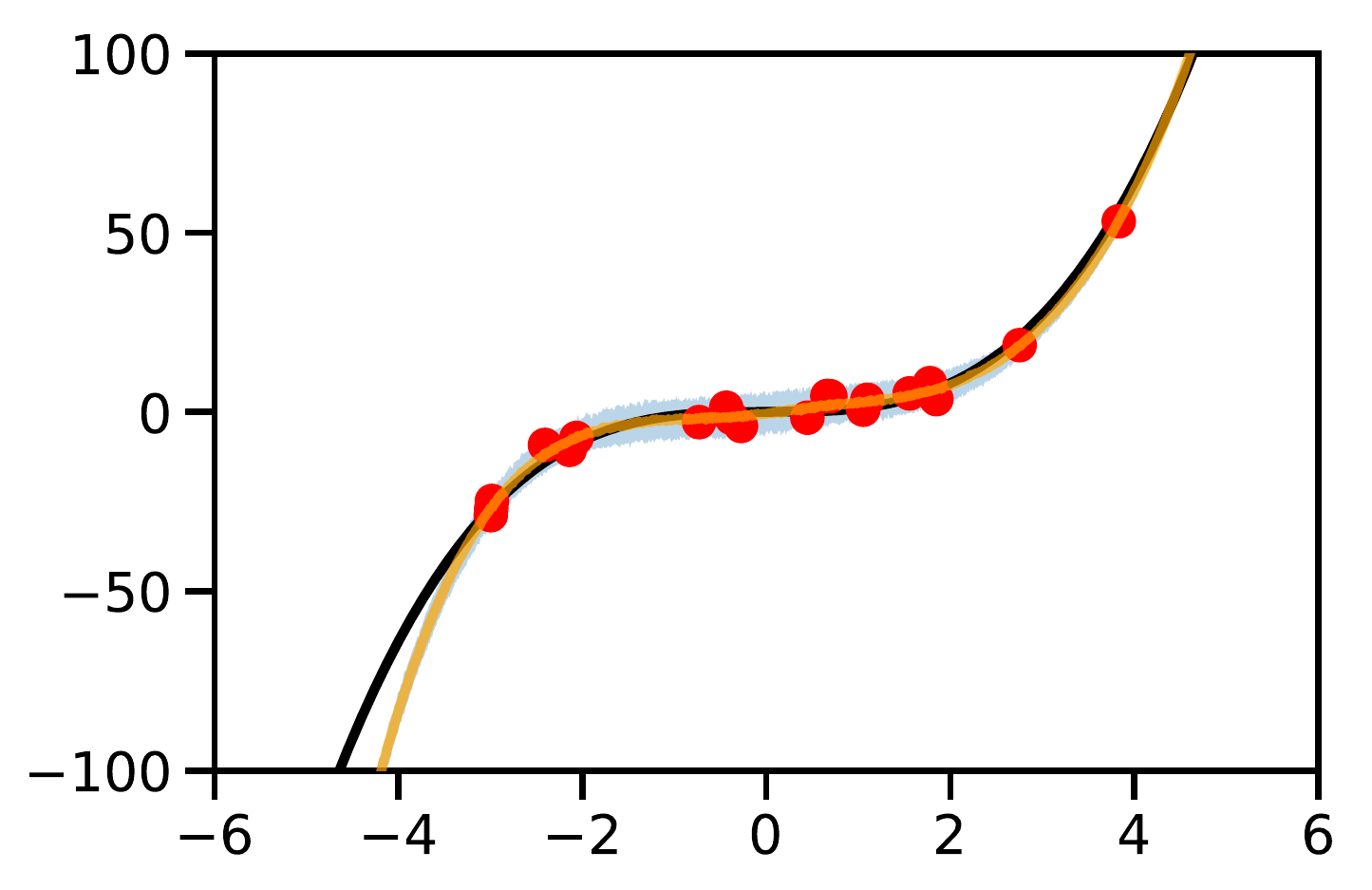}}
    \subfloat[VB-CDN]{\includegraphics[width=0.16\linewidth]{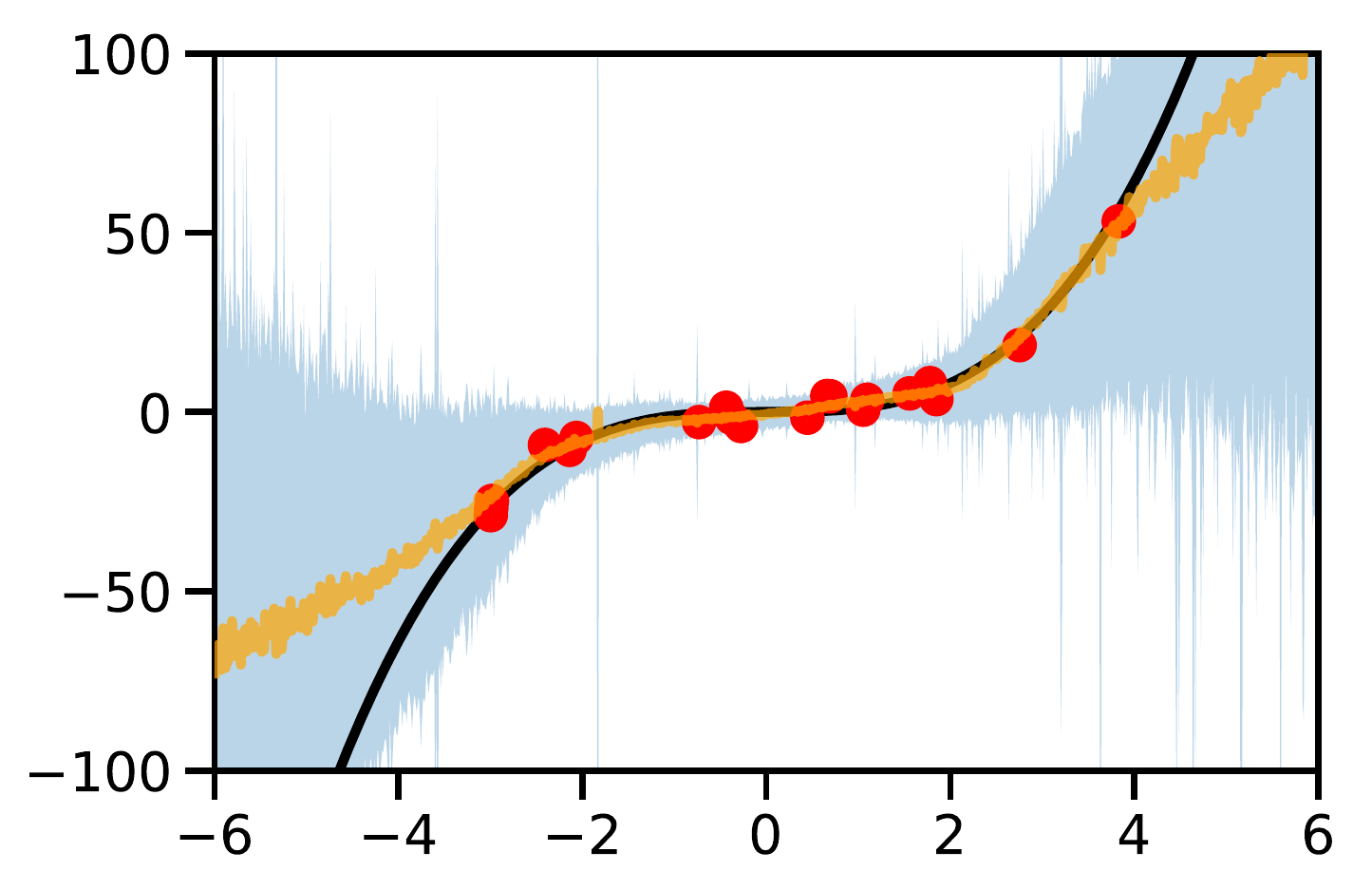}}
    \subfloat[noisy K-FAC]{\includegraphics[width=0.16\linewidth]{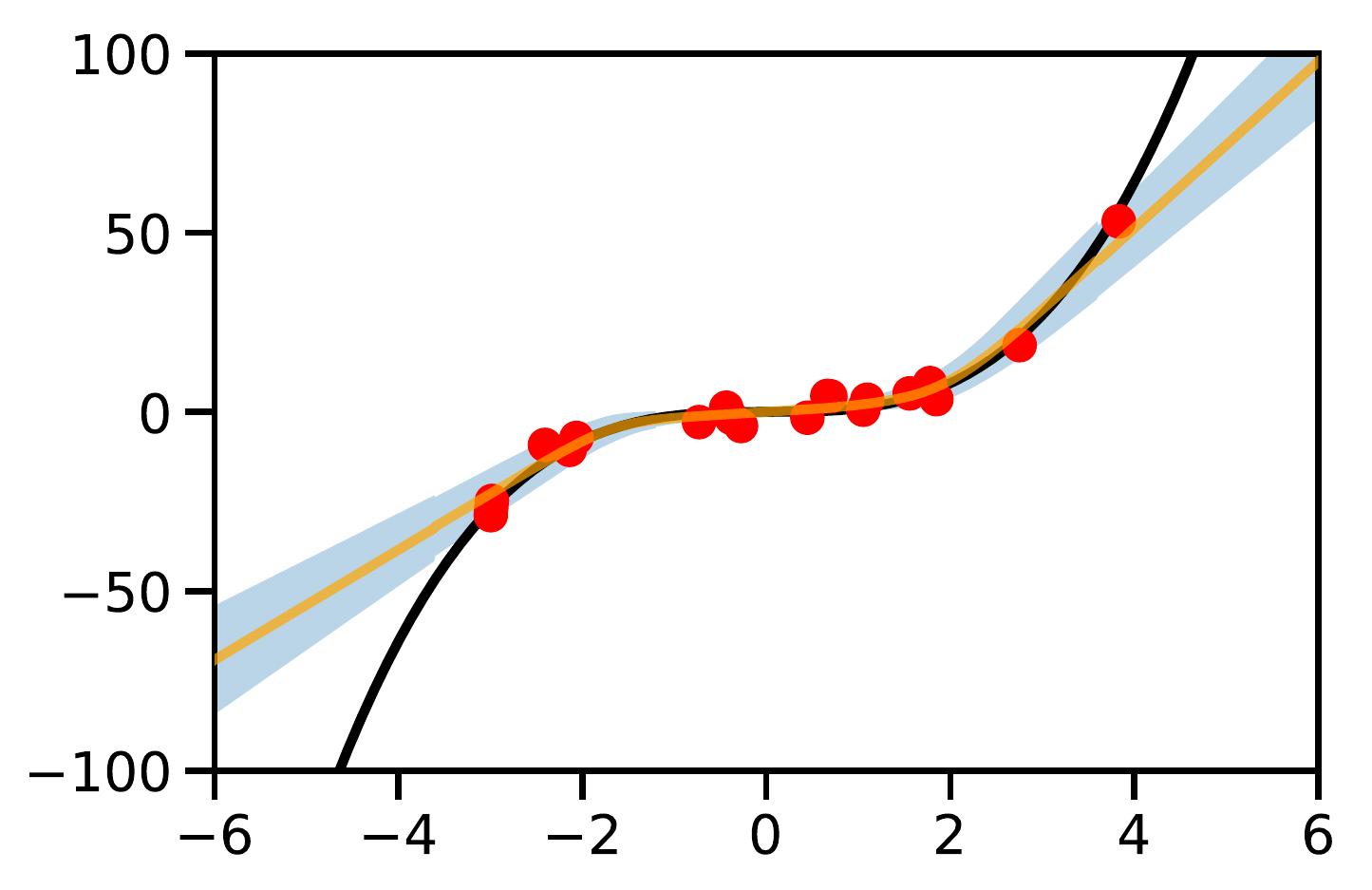}}
    \subfloat[VMG]{\includegraphics[width=0.16\linewidth]{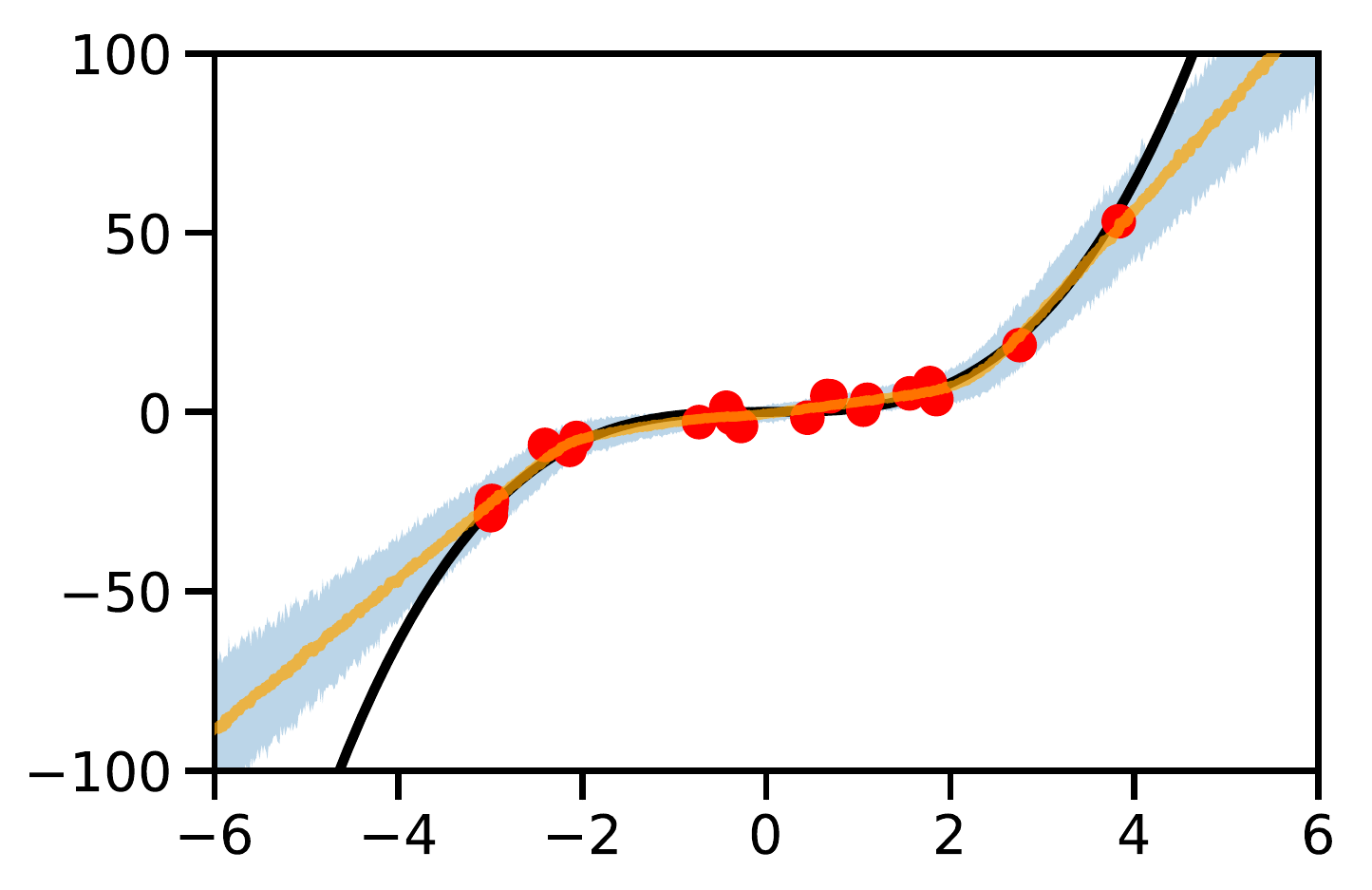}}
    \subfloat[Deep Ensemble]{\includegraphics[width=0.16\linewidth]{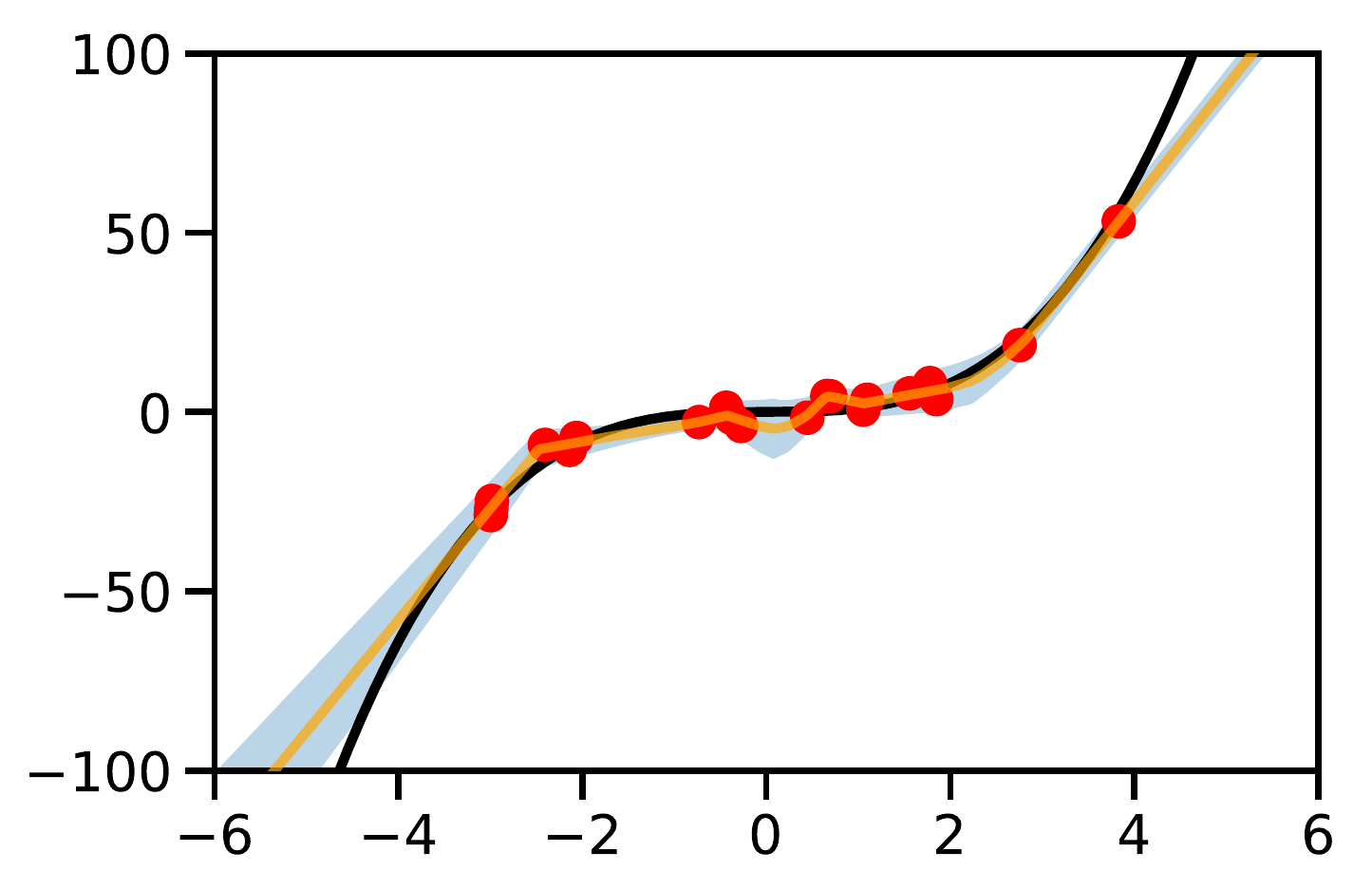}}
    \subfloat[MC-dropout]{\includegraphics[width=0.16\linewidth]{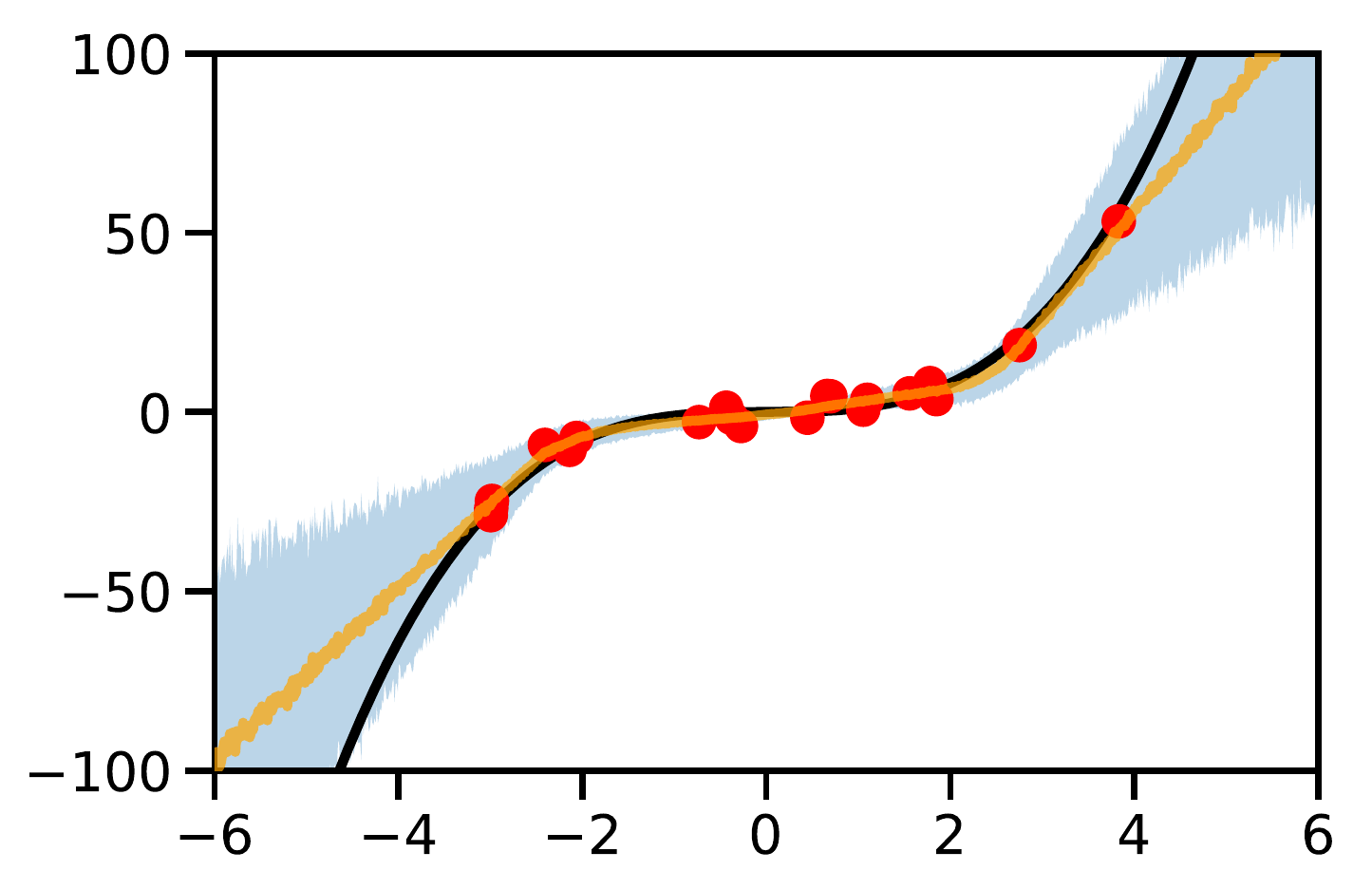}}
    
    \subfloat[ML-CDN]{\includegraphics[width=0.16\linewidth]{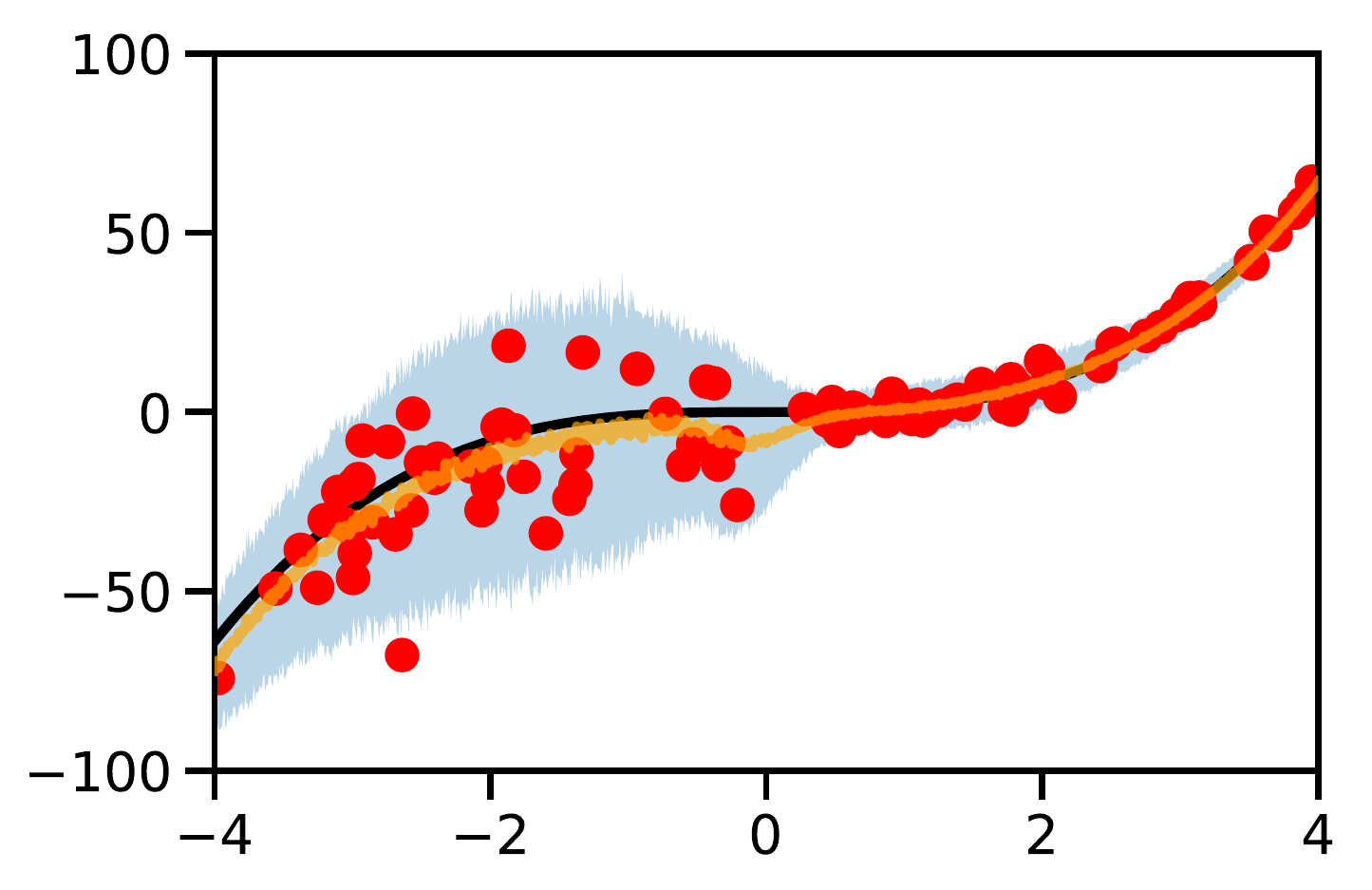}}
    \subfloat[VB-CDN]{\includegraphics[width=0.16\linewidth]{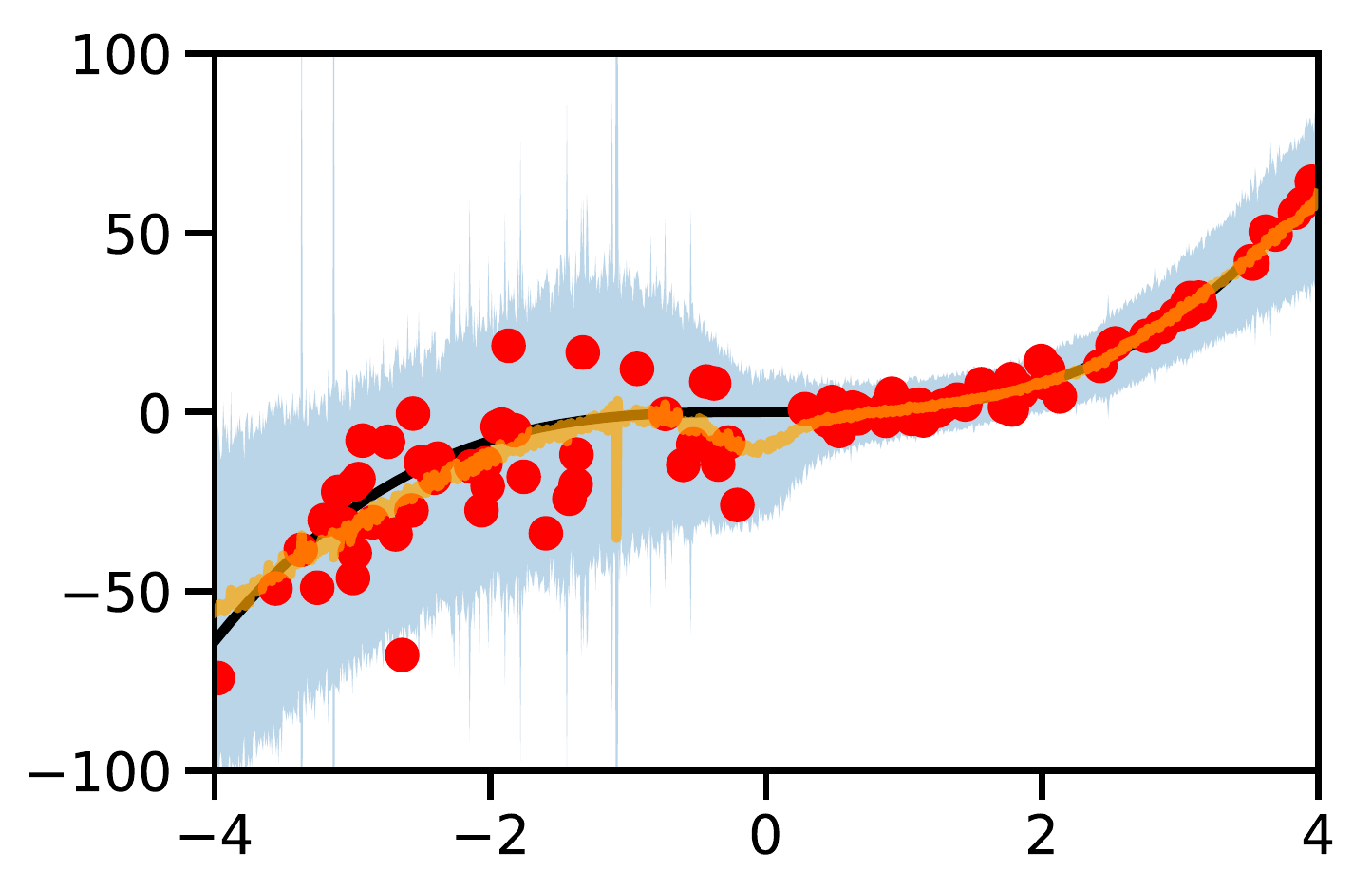}}
    \subfloat[Noisy K-FAC]{\includegraphics[width=0.16\linewidth]{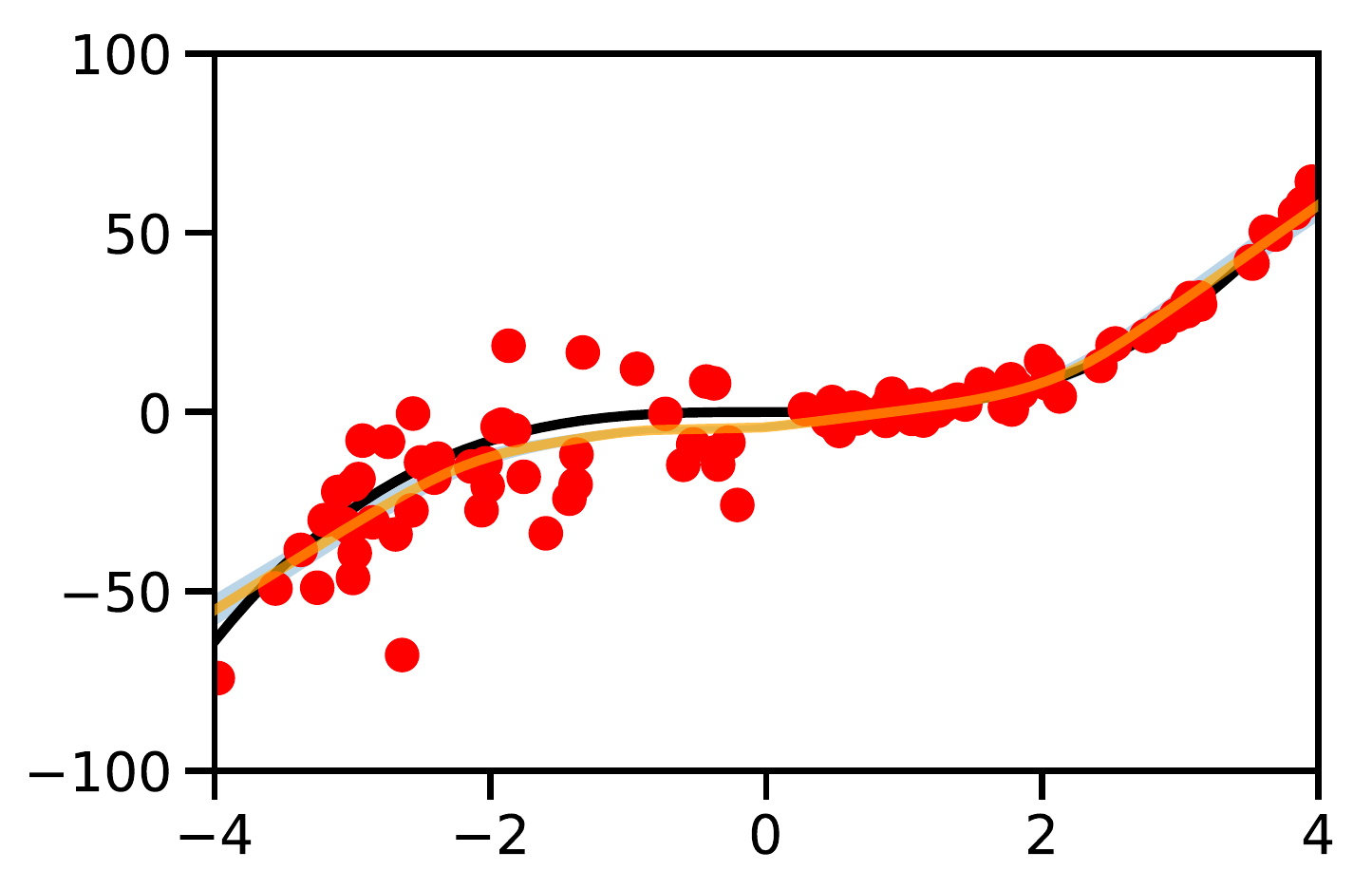}}
    \subfloat[VMG]{\includegraphics[width=0.16\linewidth]{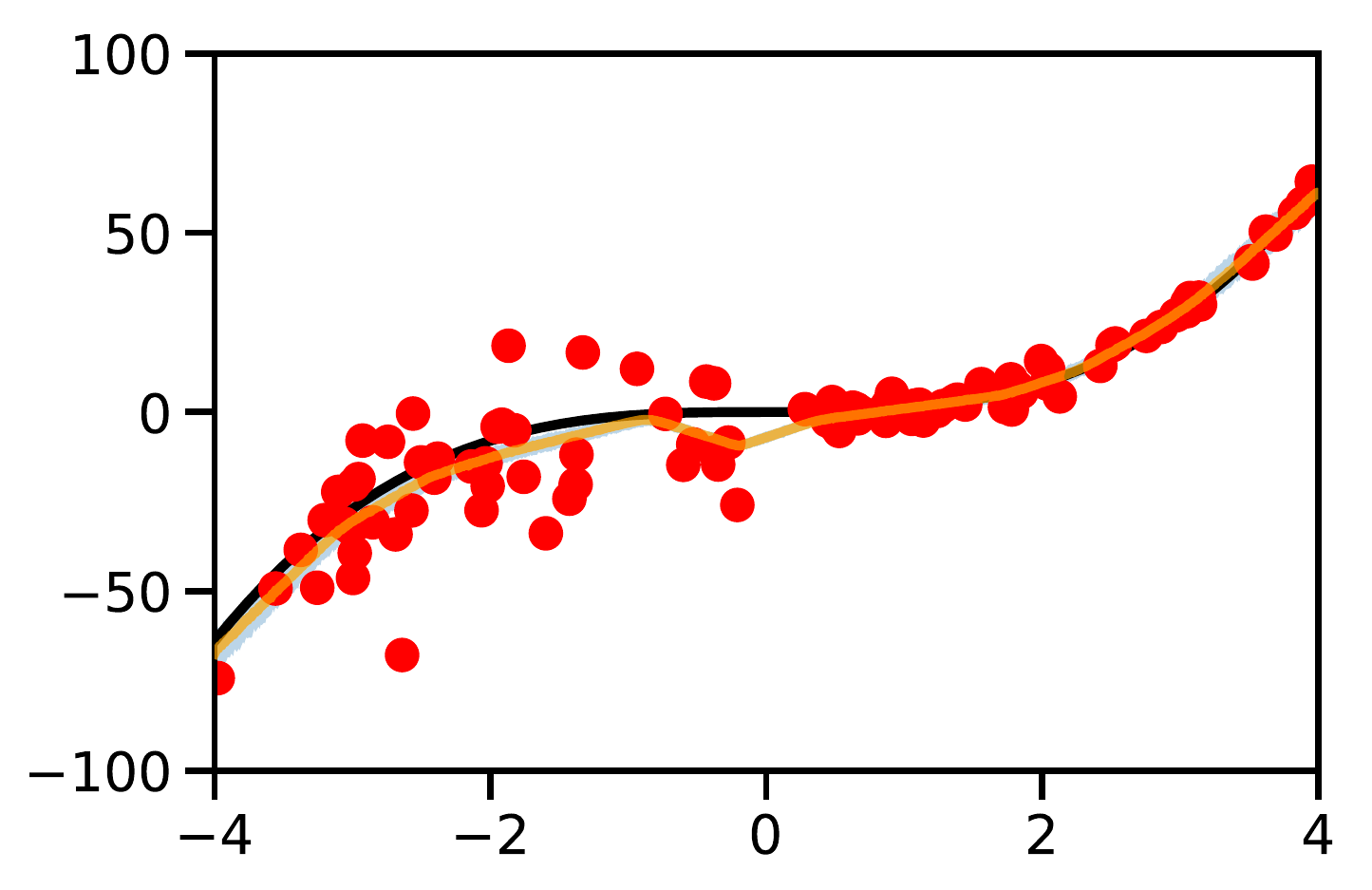}}
    \subfloat[DeepEnsemble]{\includegraphics[width=0.16\linewidth]{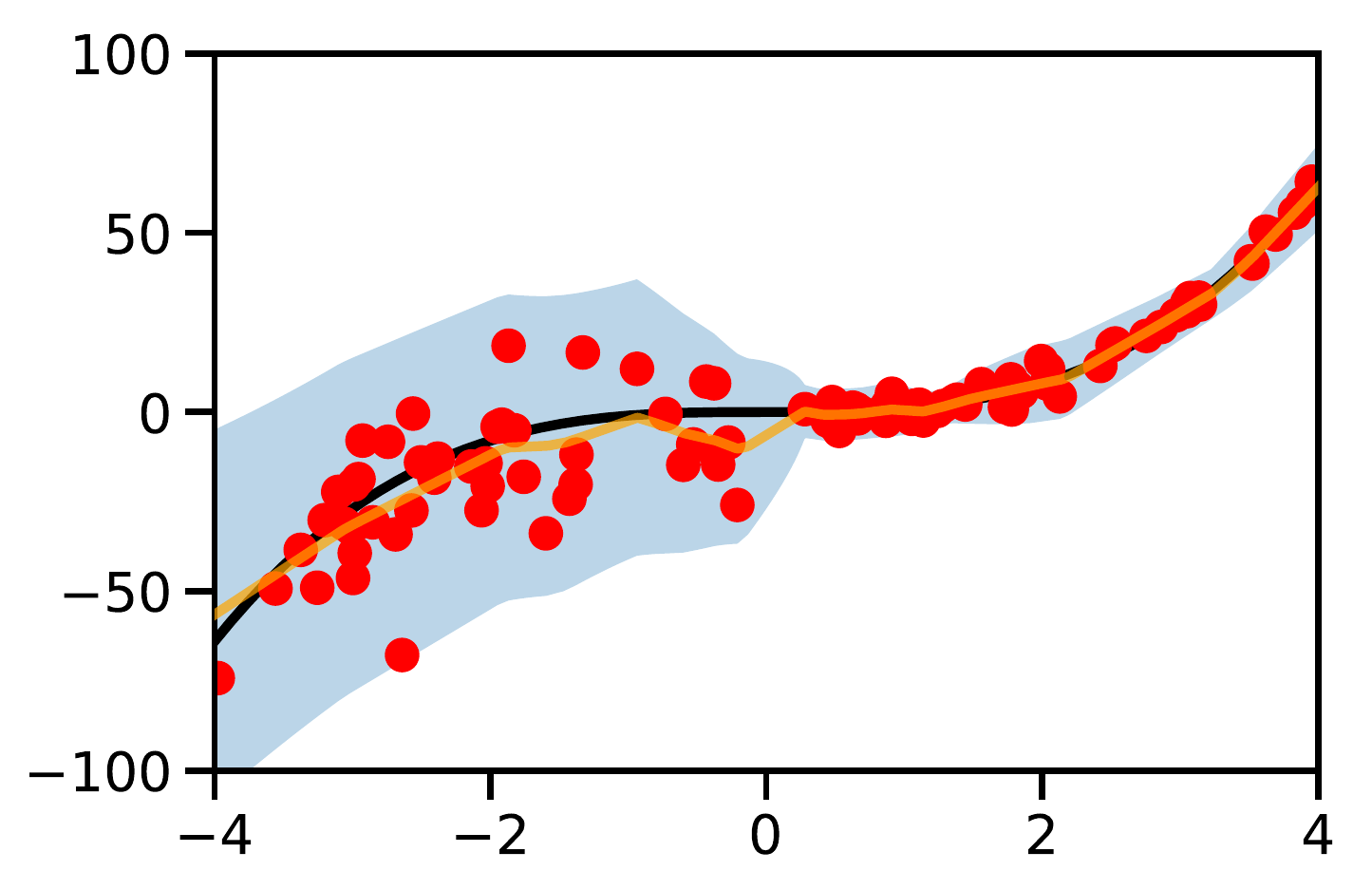}}
    \subfloat[MC-dropout]{\includegraphics[width=0.16\linewidth]{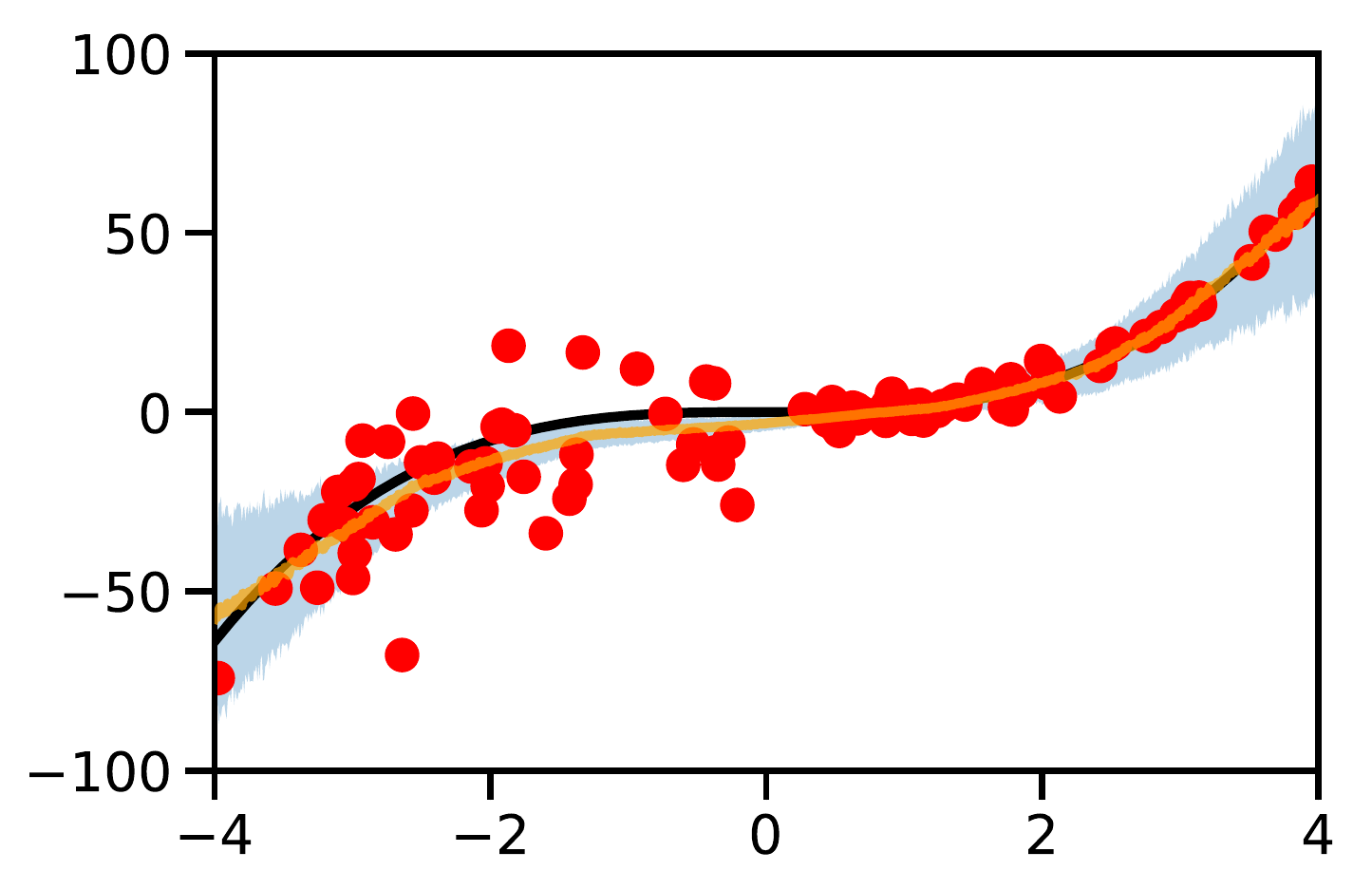}}
    
    \caption[Toy regression experiments.]{Comparison of the predictive distributions
    given by the CDNs and the baselines on toy datasets with  homoscedastic noise and few samples (first row) and heteroscedastic  noise and more samples (second row). Black lines correspond to the true noiseless function, red dots correspond to samples, orange lines and shaded regions correspond to the empirical mean and the $\pm$3 standard deviation of the predictive distribution, respectively. The VB-CDN is the only model able to capture both the epistemic uncertainty on the first and the aleatoric on the second dataset.}
    \label{fig:toy_reg}
\end{figure*}

\section{Related work}
\label{sec:related}

Various approaches for quantifying predictive uncertainty in NNs have been
proposed. Applying Bayesian inference to NNs, i.e.~treating the network parameters as random variables and estimating the posterior distribution given the training data based on Bayes' theorem, results in BNNs\citep[etc]{mackay1992practical, neal1995bayesian, graves_practical_2011, blundell_weight_2015, louizos_structured_2016, sun_learning_2017, louizos2017multiplicative,ritter_scalable_2018, DBLP:journals/corr/abs-1712-02390}. Since the true posterior distribution is intractable, BNNs are trained based on approximate inference methods such as variational inference (VI)
\citep{peterson1987mean, hinton1993keeping, graves_practical_2011, blundell_weight_2015}, Markov Chain Monte Carlo \citep{neal1995bayesian}, or Laplace approximation \citep{mackay1992practical, ritter_scalable_2018}. The final prediction is then given by the expectation of the network prediction (given the parameters) w.r.t.~the approximate posterior distribution. In VI, many modeling choices for BNNs' approximate posterior have been proposed. \citet{louizos_structured_2016} proposed to train BNNs with an MVN as the approximate posterior of each weight matrix (leading to a model they refer to as VMG). 
Multiplicative normalizing flow (MNF) \citep{louizos2017multiplicative} models the approximate posterior as a compound distribution, where the mixing density is given by a normalizing flow. \citet{DBLP:journals/corr/abs-1712-02390} also use an MVN approximate posterior and apply approximate natural gradient  \citep{amari1998natural} based maximization on the VI objective, which results in an algorithm called noisy K-FAC. 
Meanwhile, the Kronecker-factored Laplace approximation (KFLA) \citep{ritter_scalable_2018} extends the classical Laplace approximation by using an MVN approximate posterior with tractable and efficiently computed covariance factors, based on the Fisher information matrix.

Models with similar functional form as 
CDNs, i.e.~$p(\vy \vert \vx; \vpsi) = \int p(\vy \vert \vx; \vtheta) p(\vtheta \vert \vx; \vpsi) \, \dint{\vtheta}$, have been previously studied in various settings. The deep variational information bottleneck (VIB) \citep{alemi2017deep} 
assumes $\vtheta$ 
to be the hidden units of a certain layer instead of the parameters of an NN 
and trains the model with an objective derived from the information bottleneck method \citep{tishby2000information}. Interestingly, this objective gets equivalent to the ML objective for CDNs when approximated based on a single sample of $\vtheta$.
\citet{malinin2018predictive} proposed Prior Networks (PNs) which can be described by $p(\vy \vert \vx) = \int p(\vy; \vtheta) p(\vtheta \vert \vx; \vpsi) \, \dint{\vtheta}$ and aim at modeling data uncertainty with the component distribution and 
what they call ``distributional uncertainty'' (i.e.~uncertainty due to mismatch between the distributions of test and training data) with the mixture distribution. Specifically, they propose Dirichlet Prior Networks (DPNs) for uncertainty quantification in classification tasks, where $p(\vtheta \vert \vx; \vpsi)$ is assumed to be Dirichlet distribution. In contrast to CDNs, DPNs use only a single NN to parametrize the model
and an objective that augments likelihood maximization/KL-divergence minimization by a term  explicitly making use of out-of-distribution samples.
\citet{depeweg2017learning,depeweg2018decomposition} investigated BNNs with latent variables $\vtheta$ (referred to as BNN+LV), which can be described by $p(\vy \vert \vx) = \iint p(\vy \vert \vx, \vtheta; \vpsi) p(\vtheta \vert \vx) p(\vpsi) \, \dint{\vtheta} \, \dint{\vpsi}$. This model is similar to 
VB-CDNs, but in contrast 
assumes that $\vtheta$ is independent of $\vpsi$, 
which serves as additional input to the component distribution instead.
Furthermore, BNN+LV employ an $\alpha$-divergence-based objective, instead of the ELBO.

There have been several concurrent works \citep{krueger_bayesian_2017,louizos2017multiplicative,pawlowski_implicit_2017,sheikh2017stochastic} applying hypernetworks~\citep{jia2016dynamic,ha_hypernetworks_2016} to model the posterior distribution over network parameters in BNNs. 
\citet{krueger_bayesian_2017} and \citet{louizos2017multiplicative} use normalizing flows, while \citet{pawlowski_implicit_2017} and \citet{sheikh2017stochastic} use arbitrary NNs as their hypernetworks.
Note, that in Bayesian CDNs the hypernetworks themselves become BNNs.

\citet{gal2016dropout} developed  a theoretical framework that relates dropout training in NNs to approximate Bayesian inference and, as a result, proposed to approximate the predictive distribution by an average over the different networks resulting from independently sampled dropout-masks, a technique which they referred to as MC-dropout (MCD) and which they applied to estimate the prediction uncertainty in NNs. Recently,  \citet{lakshminarayanan2017simple}, proposed to use an ensemble of NNs in conjunction with a proper scoring rule and adversarial training to quantify the prediction uncertainty of deep NNs, leading to a model referred to as Deep Ensemble (DE). The DE provides a  non-Bayesian way to quantify prediction uncertainty, and is in this sense related to the  approaches of \citet{guo17calibration} and \citet{hendrycks17baseline}.

\section{Experiments}
\label{sec:experiment}

We consider several standard tasks in our experimental analysis: 1D toy regression problems inspired by~\citet{hernandez2015probabilistic} (\Cref{subsec:toy_reg}), classification under out-of-distribution data (\Cref{subsec:ood_clf}), and detection of and defense against adversarial examples~\citep{szegedy2013intriguing} (\Cref{subsec:adv_ex}).
We refer to the CDNs that are trained via $\mathcal{L}_\text{ML}$ (\cref{eq:MLobjective}) and $\mathcal{L}_\text{VB}$ (\cref{eq:CDN_ELBO}) as ML-CDNs and VB-CDNs, respectively.
The following recent models (described in \Cref{sec:related}) are considered as the baselines: VMG, MNF, DPN, noisy K-FAC, MC-dropout, and Deep Ensemble.\footnote{Additional baselines are investigated in the supplement.}


We estimate the predictive distribution $p(\vy \vert \vx)$ of the CDNs, based on 100 joint samples of $
\vpsi \sim q(\vpsi; \vomg), \vtheta \sim p(\vtheta; g(\vx; \vpsi))$ for VB-CDNs and 100 samples of $\vtheta \sim p(\vtheta; g(\vx; \vpsi))$ for ML-CDNs. We also draw 100 samples from the posterior to approximate the predictive distribution of BNN baselines.
If not stated otherwise, we use a single sample to perform Monte Carlo integration during training. 
We pick the regularizaion hyperparameter $\lambda$ for ML-CDNs (\cref{eq:MLobjective}) out of the set $\{ 10^{-4}, 10^{-5}, 10^{-6}, 10^{-7}, 10^{-8} \}$ which maximizes the validation accuracy.
We use Adam~\citep{kingma2014adam} with default hyperparameters for optimization in all experiments and the implementations provided by \citet{louizos2017multiplicative}\footnote{\url{https://github.com/AMLab-Amsterdam/MNF_VBNN}} and \citet{DBLP:journals/corr/abs-1712-02390}\footnote{\url{https://github.com/gd-zhang/noisy-K-FAC}} for MNF and noisy K-FAC, respectively.
Where mini-batching is necessary, e.g.~on MNIST and CIFAR-10, we use mini-batches of size 200. 
All models are optimized over 10000 iterations in  the toy regression experiments, 20000 iterations ($\approx$67 epochs) in experiments on MNIST and Fashion-MNIST, and 100 epochs in experiments on CIFAR-10. We chose ReLU and hyperbolic tangent as the nonlinearity of the ML-CDNs' and VB-CDNs' hypernetworks, respectively. For the details on the selection of the model-specific hyperparameters of the baselines, we refer the reader to the supplementary material. The source code for all our experiments is available at 
\url{https://anonymous.com}.

\begin{figure*}[t!]
    \centering
    \subfloat[MNIST]{\includegraphics[width=0.235\linewidth]{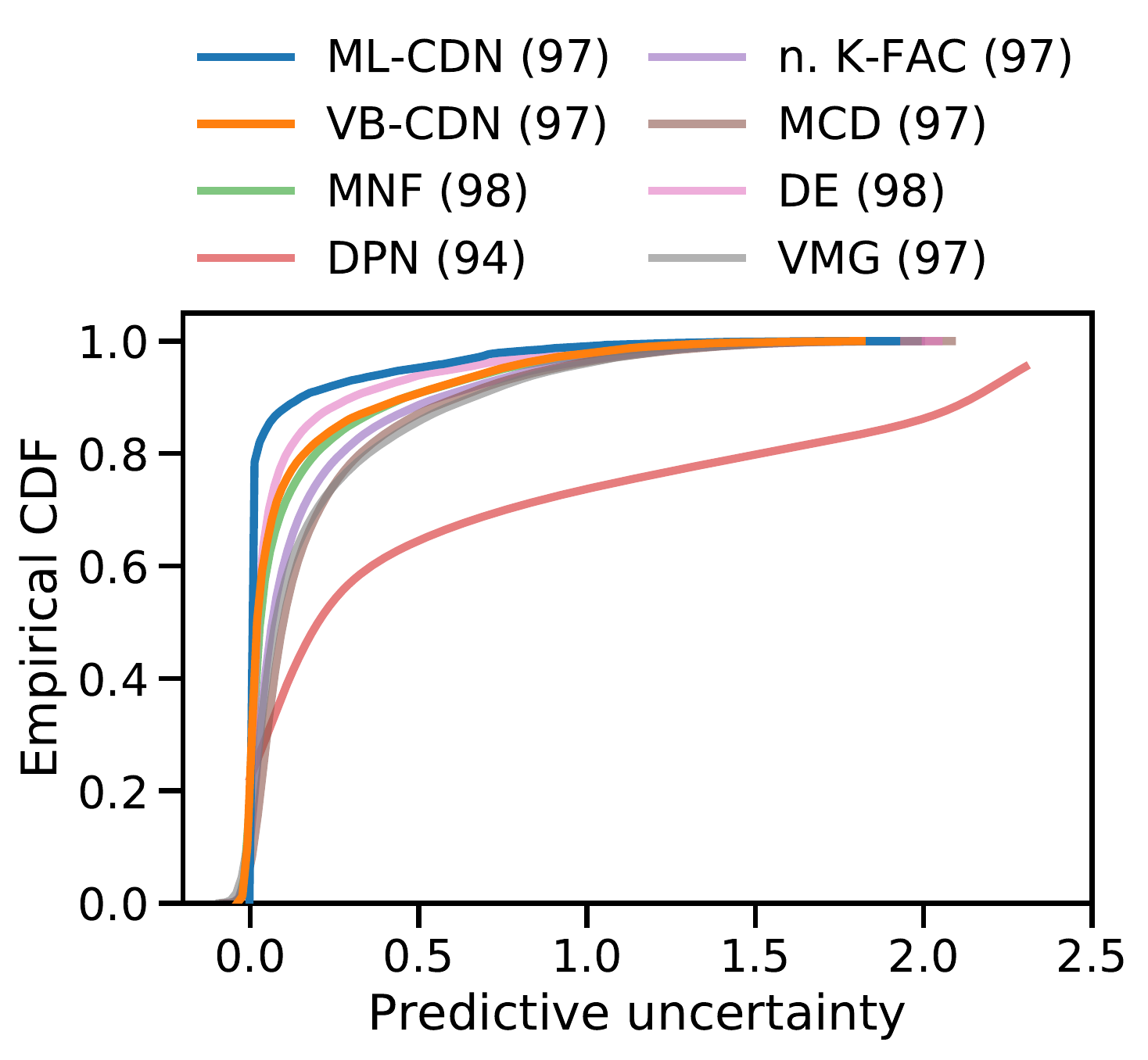}}
    \subfloat[notMNIST]{\includegraphics[width=0.235\linewidth]{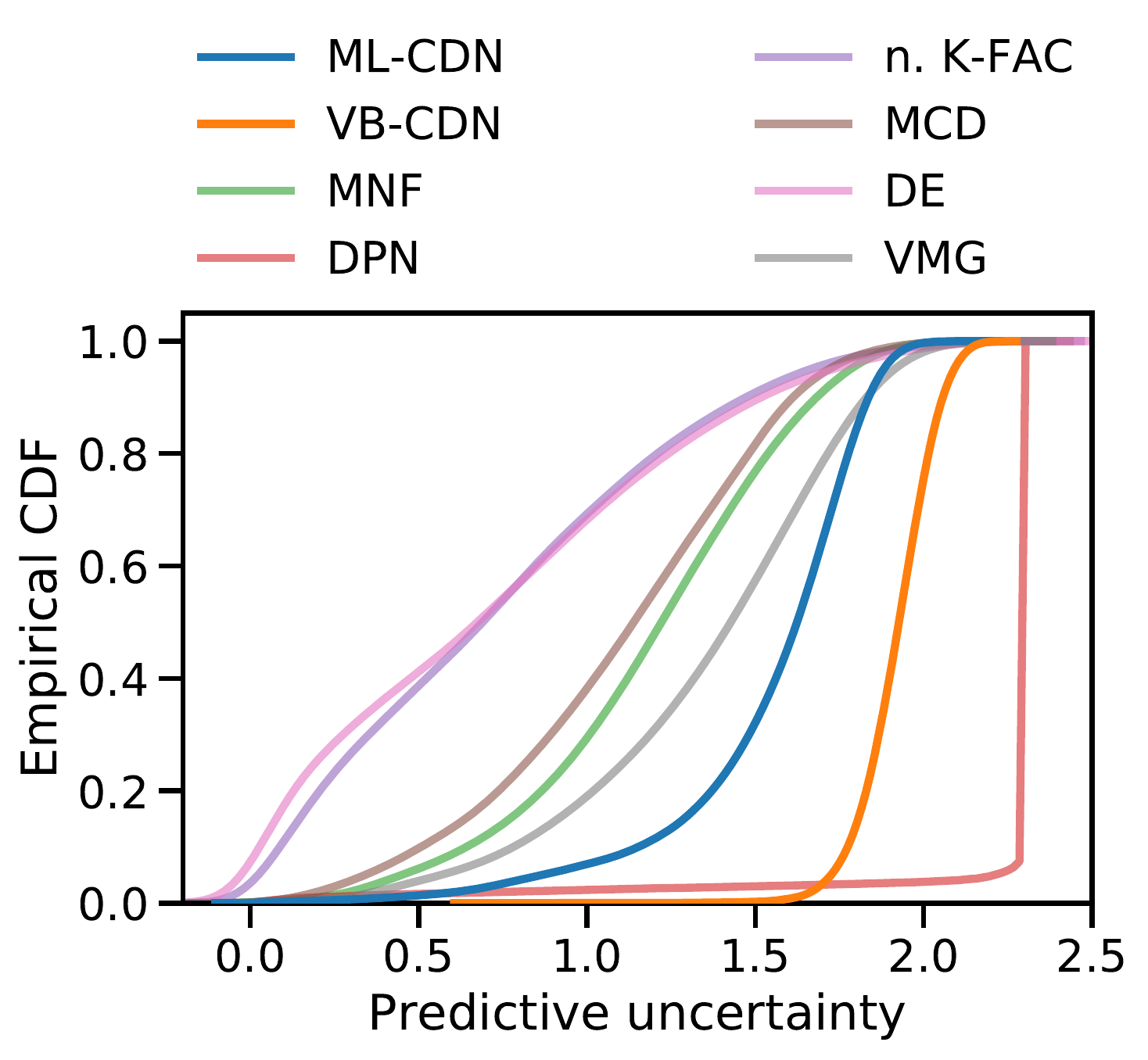}}
    \qquad
    \subfloat[Fashion-MNIST]{\includegraphics[width=0.235\linewidth]{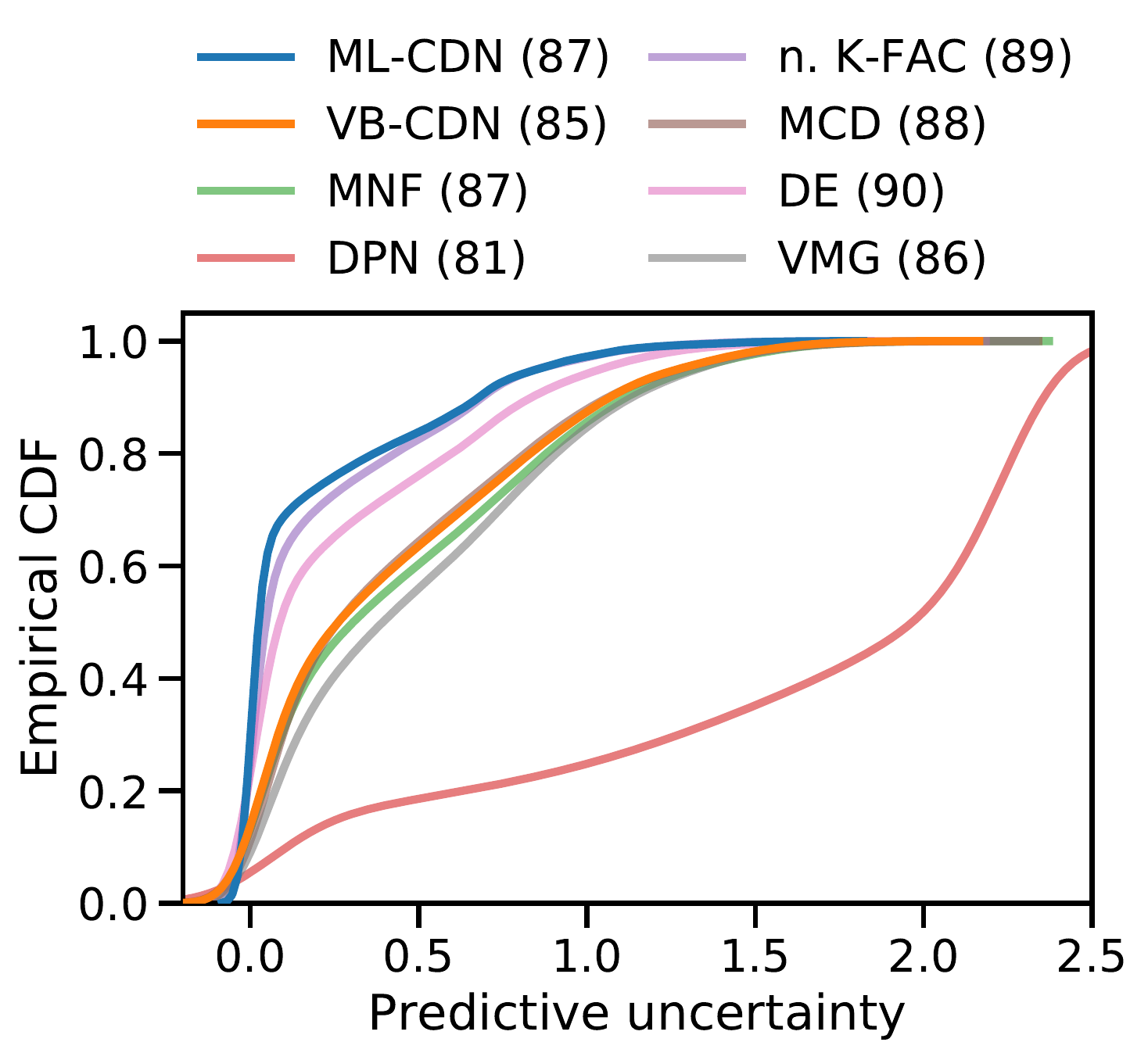}}
    \subfloat[Flipped Fashion-MNIST]{\includegraphics[width=0.235\linewidth]{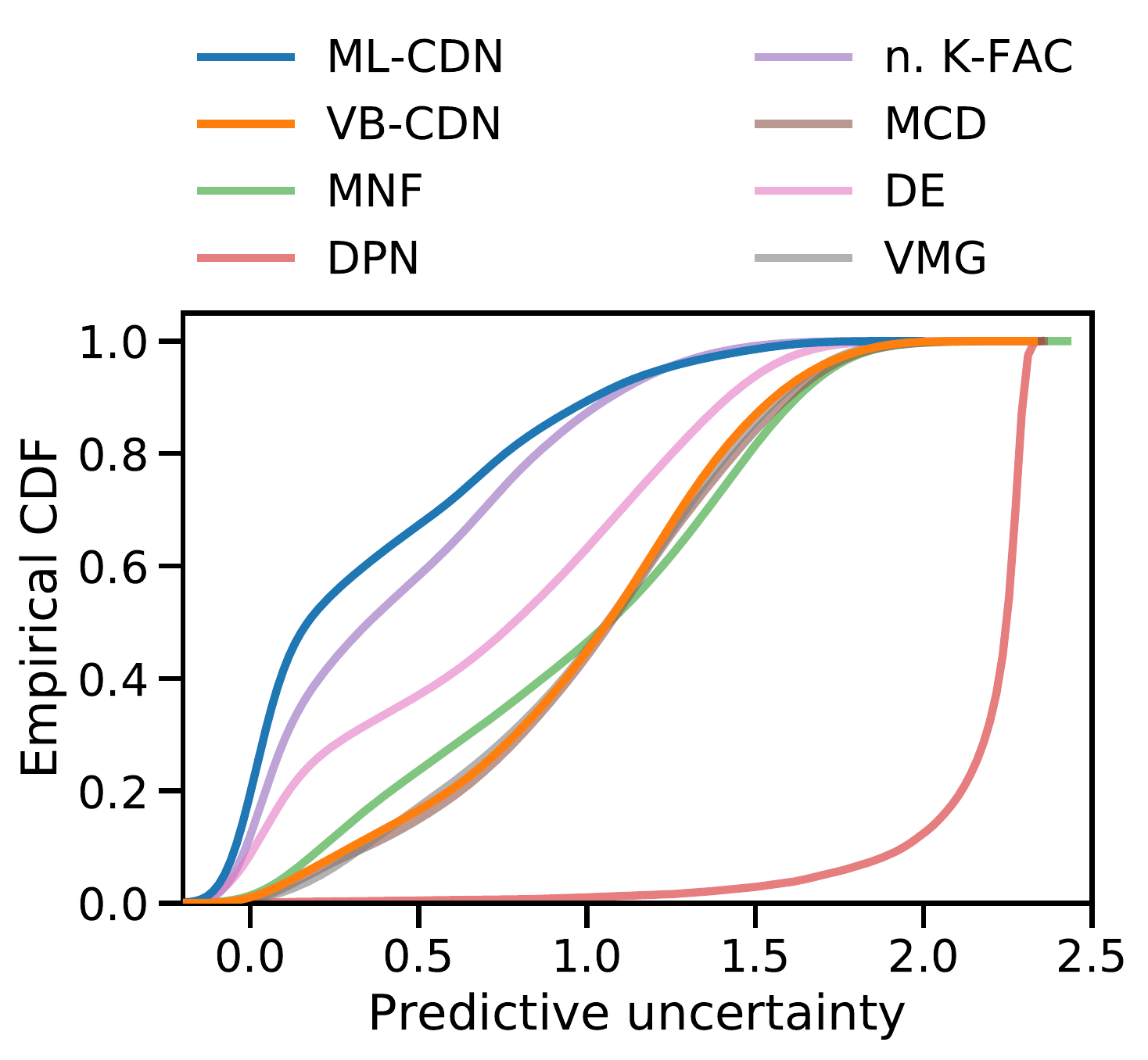}}
    
    \caption{CDFs of the empirical entropy of the predictive distribution of the models, trained on MNIST (the first two figures) and Fashion-MNIST (the last two figures). The caption of each figure indicates the test set used, the y-axis denotes the fraction of predictions having entropy less than the corresponding value on the x-axis, and the number next to each model name indicates its test accuracy, in percent.}
    \label{fig:notmnist}
\end{figure*}

\subsection{Toy regression}
\label{subsec:toy_reg}


    
    

Following \citet{hernandez2015probabilistic}, we generate the first toy regression dataset as follows: We sample 20 input points $x \sim \mathcal{U}[-4, 4]$  and their target values $y = x^3 + \epsilon$, where $\epsilon \sim \mathcal{N}(0, 3^2)$, i.e.~the data noise is homoscedastic. We aim at analyzing how well the target function is modeled over the larger interval $[-6, 6]$. Having only few data points, it is a desirable property of a model to express high (epistemic) uncertainty in regions with no or only few samples, e.g.~between $-6$ and $-4$ or $4$ and $6$. 
The second toy regression dataset is constructed by sampling 100 data points as above, this time with different scale of noise in different intervals: $\epsilon \sim \mathcal{N}(0, 3^2)$, if $x \geq 0$. and $\epsilon \sim \mathcal{N}(0, 15^2)$, otherwise. This dataset is designated for testing whether a model can capture heteroscedastic aleatoric uncertainty. 
 
In these experiments, we use a two-layer MLP with 100 hidden units as the predictive network, while the hypernetworks of the CDNs ($g_1$ and $g_2$) are modeled with two-layer MLPs with 10 hidden units each. Three samples of $\vtheta$ (along with a single sample of $\vpsi$ in the case of the VB-CDN) are used to approximate the objectives during training of both CDNs and BNNs.\footnote{On these toy datasets, we found that using more than one sample is crucial for the results of the CDNs (i.e.~results for using just one sample look similar to that of the VMG and Noisy K-FAC as can be seen in the supplement), while it does not significantly change the behaviour of the BNNs.}
A regularization hyperparameter of $\lambda = 10^{-3}$ is used for training the ML-CDNs.

The results for the first data set (shown in the first row of~\Cref{fig:toy_reg}) demonstrate that the VB-CDN is capable of capturing the epistemic uncertainty 
like other Bayesian models. This is not the case for the ML-CDN (which displays high confidence everywhere) and the DE (which captures only the uncertainty on the left side). This demonstrates the benefits of using a Bayesian approach for capturing parameter uncertainty.
On the other hand, the mixture models, i.e.~the CDNs and the DE, are the only ones able to capture the aleatoric uncertainty on the second dataset, as shown in the second row of~\Cref{fig:toy_reg}. This can be explained by the ability of CDNs and DEs to model input-dependent variance.


To further investigate the different roles in uncertainty modeling of the mixing distribution and the approximate posterior of VB-CDNs, we compare their average variance (over parameters and samples).
\footnote{We picked $1000$ evenly spaced points from $[-6, 6]$ and $[ -4, 4 ]$ respectively and approximated the means over the posterior with $100$ samples.}
On the first data set, the average variance of the mixing distribution is $0.356$ and that of the
posterior distribution is $0.916$. On the second data set the average variance of the posterior distribution is $0.429$ and that of the mixing distribution is $0.618$ for $x<0$ and  $0.031$ for $x\geq0$. Therefore, the variance of the posterior is reduced on the second data set (as desired for more training data) while the mixing distribution successfully captures the higher data uncertainty for $x<0$, indicating that the approximate posterior successfully models epistemic and the mixing distribution aleatoric uncertainty.

\subsection{Out-of-distribution classification}
\label{subsec:ood_clf}

Following \citet{lakshminarayanan2017simple}, we train all models on the MNIST training set\footnote{We use Fashion-MNIST as OOD data for training the DPN.} and investigate their performance on the MNIST test set and the notMNIST dataset\footnote{\url{http://yaroslavvb.blogspot.com/2011/09/notmnist-dataset.html}.}, which contains images (of the same size and format as MNIST) of letters from the alphabet instead of handwritten digits. On such an out-of-distribution (OOD) test set, the predictive distribution of an ideal model should have maximum entropy, i.e.~it should have a value of $\ln 10 \approx 2.303$ which would be achieved if all ten classes are equally probable. The predictive NN used for this experiment is an MLP with a 784-100-10 architecture.

We present the results in \Cref{fig:notmnist}a and b, where we plotted the cumulative distribution function (CDF) of the empirical entropy of the predictive distribution, following \citet{louizos2017multiplicative}. A CDF curve close to the top-left corner of the figure implies that the model yields mostly low entropy predictions, indicating that the model is very confident. While one wishes to observe high confidence on data points similar to those seen during training, the model should express uncertainty when exposed to 
OOD data. That is, we prefer a model to have a CDF curve closer to the bottom-right corner on notMNIST, as this implies it makes mostly uncertain (high entropy) predictions, and a curve closer to the upper-left corner for MNIST. As the results show, the VB-CDN yields high confidence on the test set of MNIST while having significantly lower confidence on notMNIST compared to all baseline models, except the DPN. Note however, that training DPNs requires additional data (which makes the comparison unfair) and that the DPN's prediction accuracy and confidence on the MNIST test set are low compared to all other models. 
For the ML-CDN, we observe that it is more confident than all other models on within-distribution data, at the expense of showing lower uncertainty on OOD data than the VB-CDN.

Quantitaively, we calculated 
the mean maximal confidence for in-distribution (MMC-in) and OOD data (MMC-out) as well as the area under receiver operating characteristic (AUROC). The results can be found in Table \ref{sample-table}. 

\begin{table}[htb]
    \caption{Mean maximal confidence (MMC) for in distribution  (MNIST) and OOD data (notMNIST) and 
    area under receiver operating characteristic (AUROC).} 
    \begin{center}
    \begin{tabular}{lccc}
    \toprule
    \textbf{Algorithm}  & \textbf{MMC-in} & \textbf{MMC-out} &  \textbf{AUROC} \\
    \midrule 
    CDN & \textbf{0.978} & \textbf{0.430} & \textbf{0.993} \\
    VMG & 0.938 & 0.507 & 0.964 \\
    MNF & 0.959 & 0.504 & 0.977\\
    MCD & 0.950 & 0.665 & 0.928 \\
    DE  & 0.970 & 0.740 & 0.862 \\
    noisy-KFAC & 0.949 & 0.744 & 0.848 \\
    \bottomrule
    \end{tabular}
    \end{center}
    \label{sample-table}
\end{table}
Our model clearly has the highest MMC value for in-distribution data and the highest AUROC, 
while having the lowest MMC value for OOD data.
%

 On the more challenging OOD task introduced by \citet{alemi2018uncertainty} where Fashion-MNIST \citep{xiao2017/online} is used as training set\footnote{We use MNIST as OOD data for training the DPN.}, while the vanilla and the up-down flipped test set of Fashion-MNIST are used for evaluation (\Cref{fig:notmnist}c and d), the results are less pronounced, but the CDNs still show a performance competitive to that of the baseline models.

\begin{figure}[htb]
    \centering
    \subfloat[Accuracy]{\includegraphics[width=0.5\linewidth]{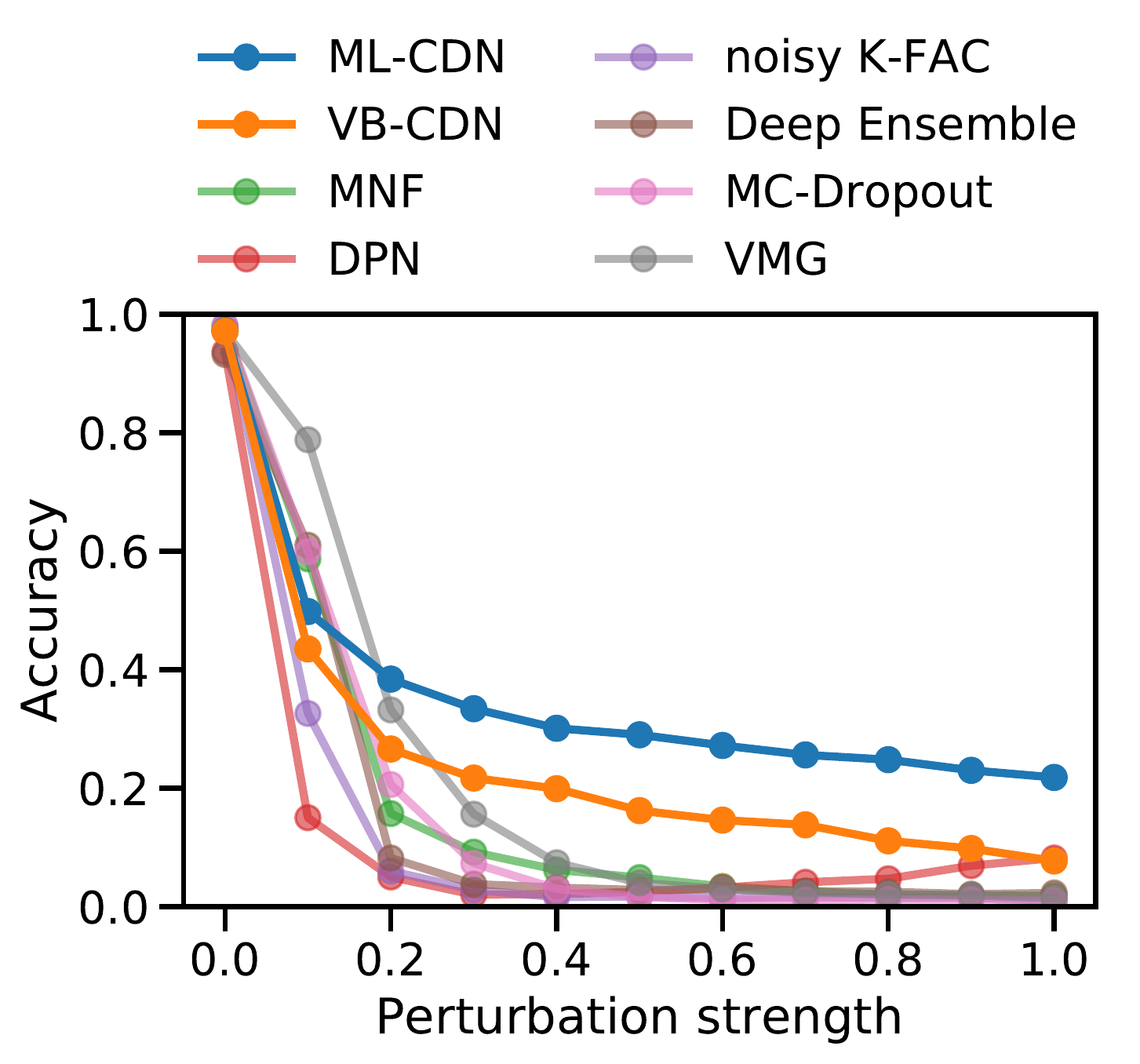}}
    \subfloat[Entropy]{\includegraphics[width=0.5\linewidth]{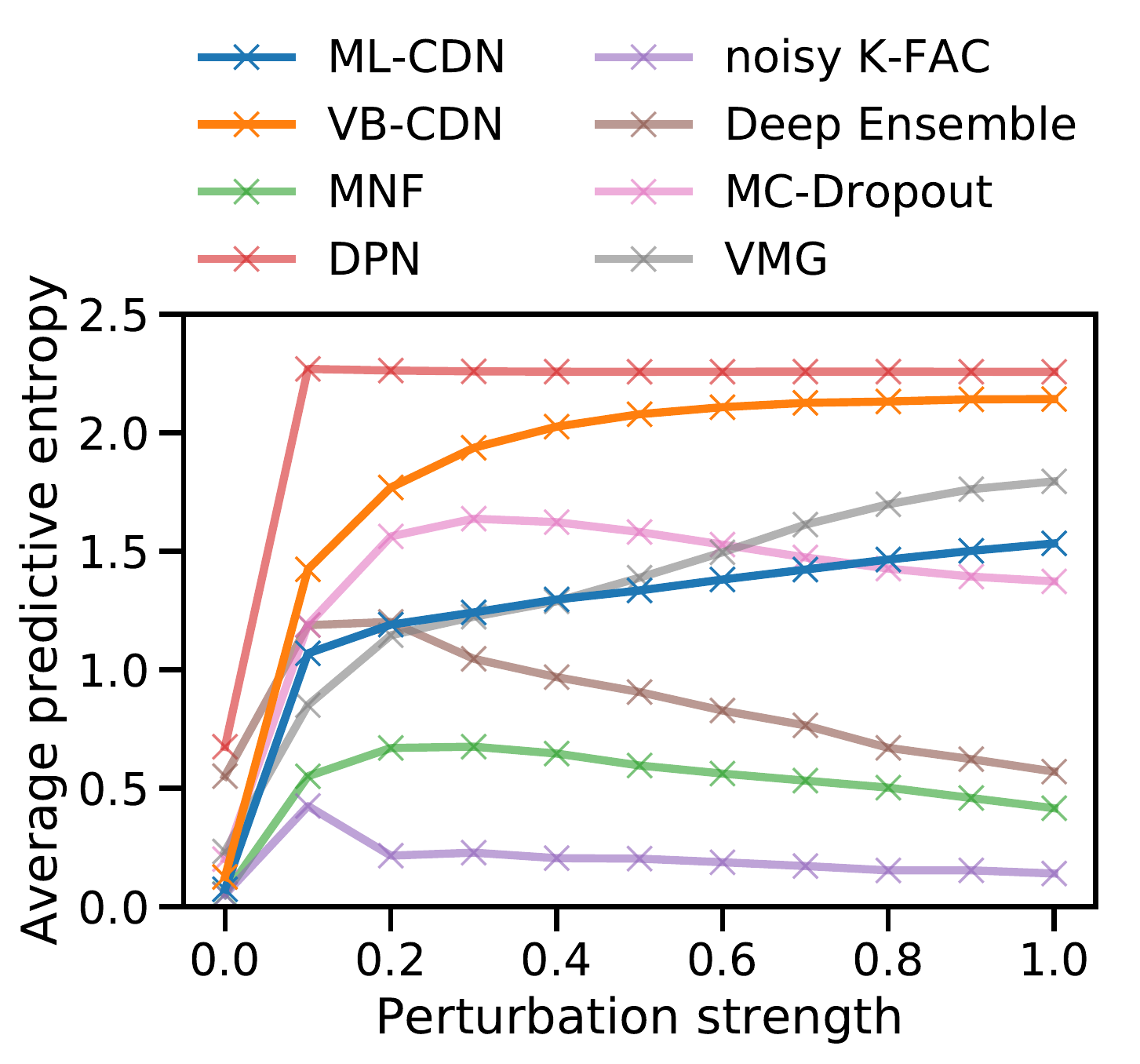}}
    
   \caption[MNIST adversarial attack experiments.]{Prediction accuracy and average entropy of models trained on MNIST when attacked by FGSM-based adversarial examples
   \citep{goodfellow6572explaining} with varying perturbation strength.}
    \label{fig:adv_examples}
\end{figure}

\begin{figure}[htb]
    \centering
    \subfloat[ML-CDN]{\includegraphics[width=0.5\linewidth]{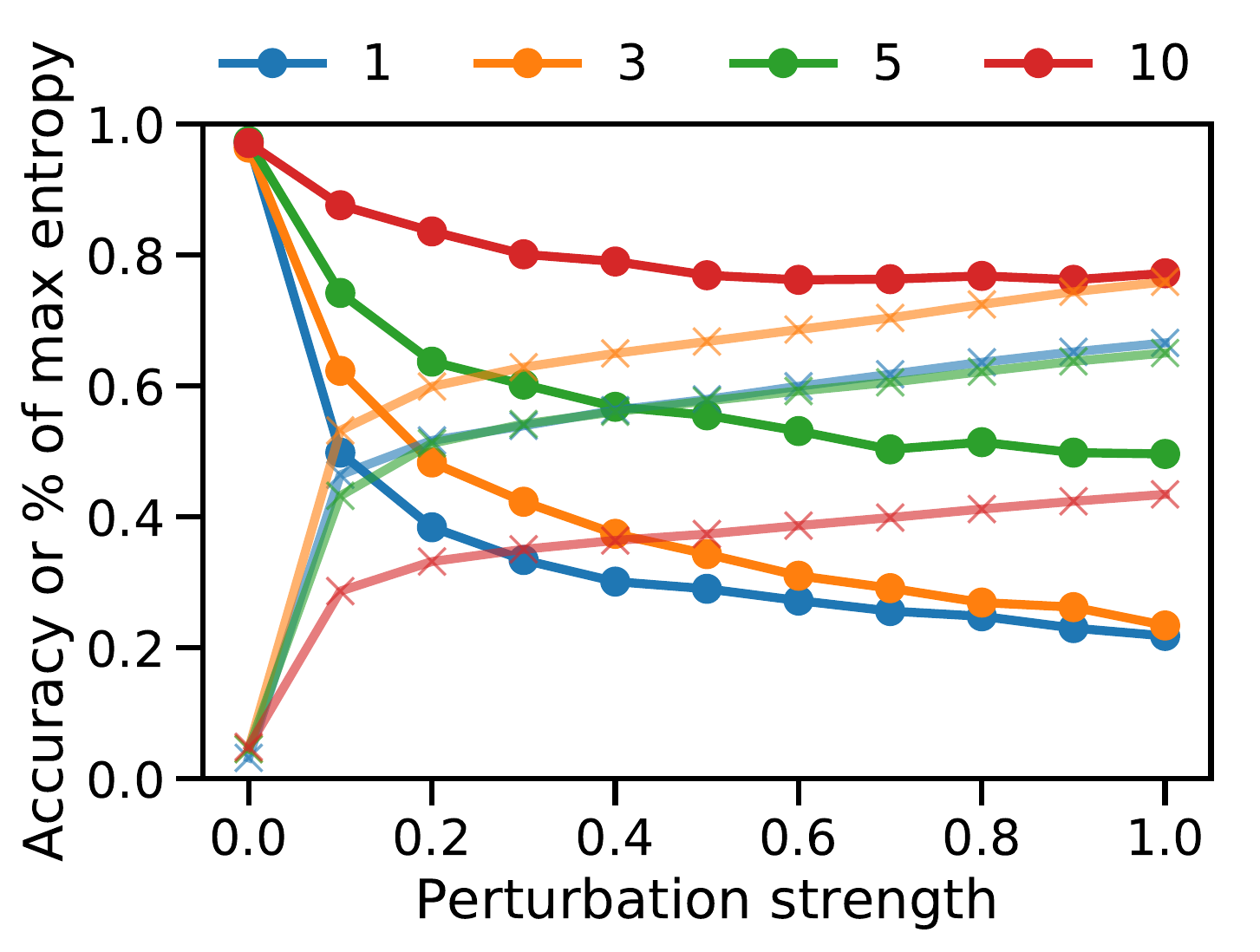}}
    \subfloat[VB-CDN]{\includegraphics[width=0.5\linewidth]{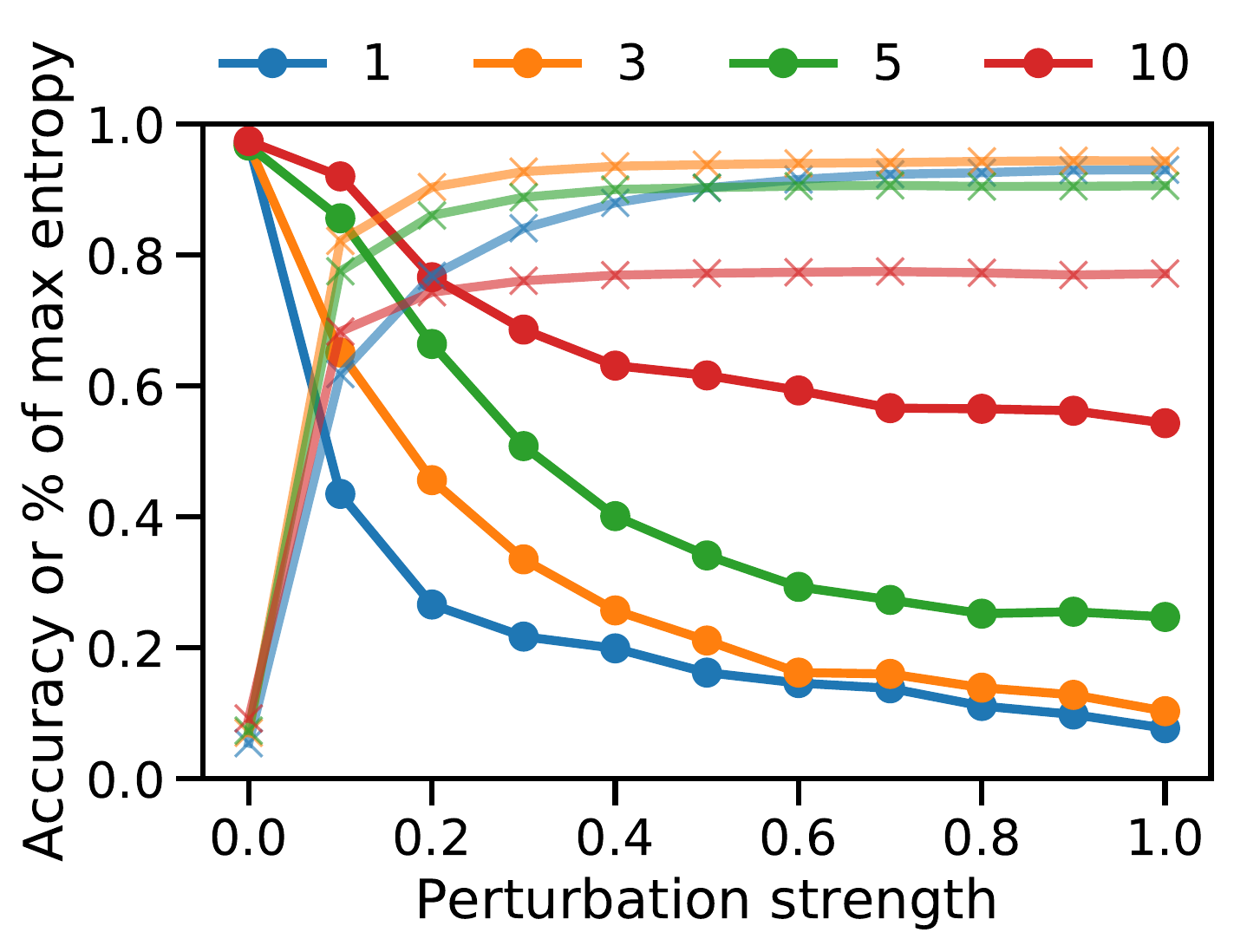}}
    
    \subfloat[BNN (Noisy K-FAC)]{\includegraphics[width=0.5\linewidth]{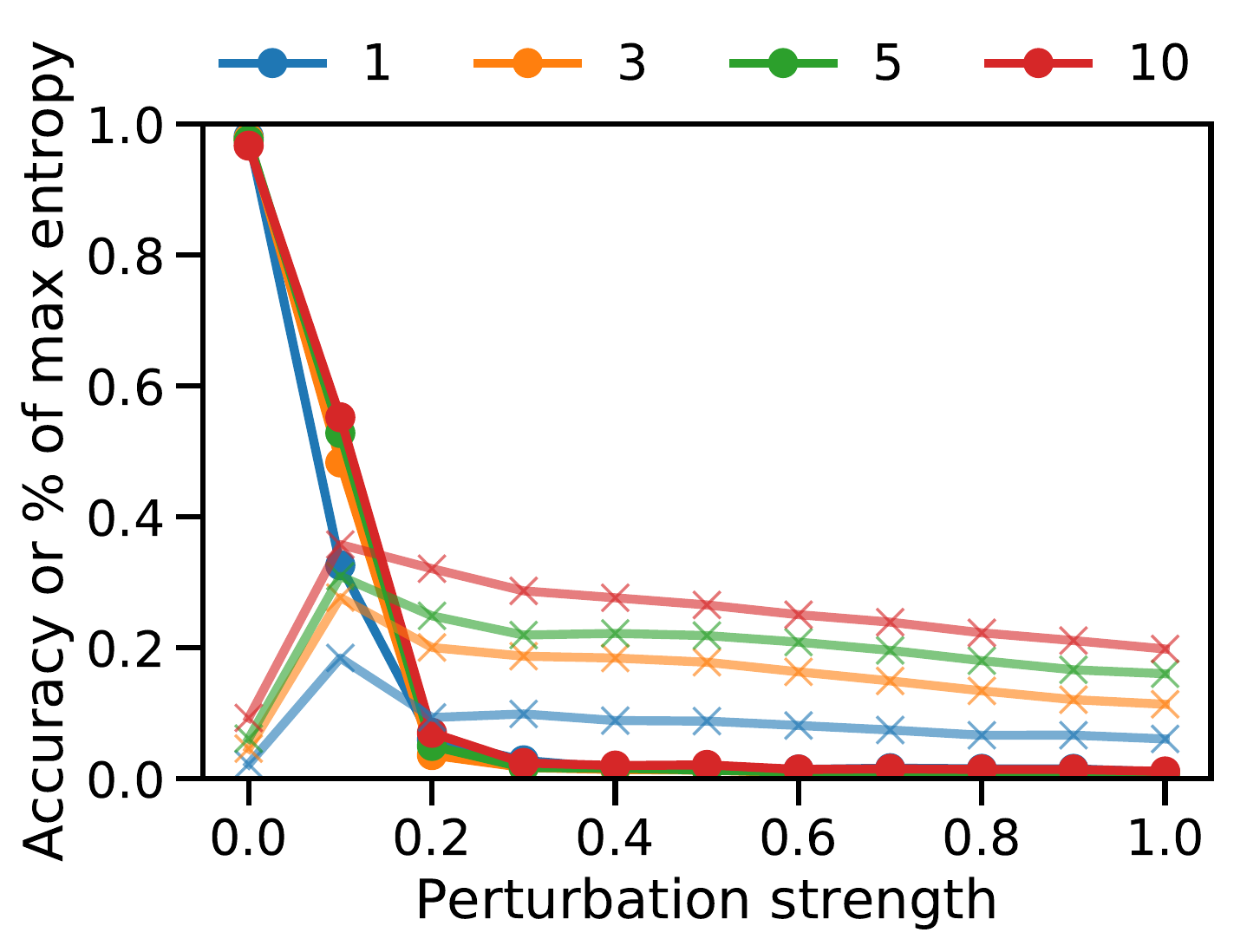}}
    \subfloat[BNN (VMG)]{\includegraphics[width=0.5\linewidth]{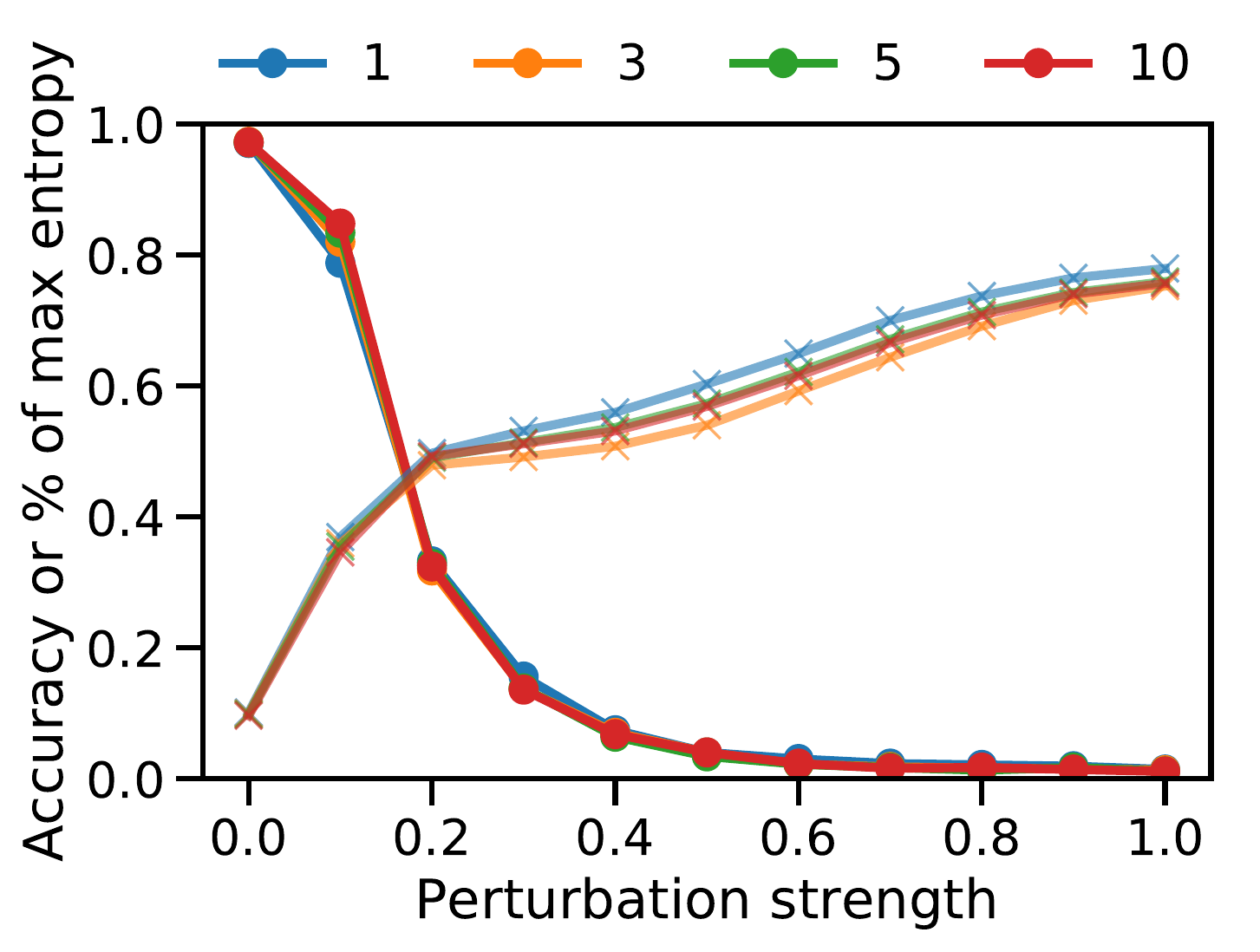}}
    
    \caption[MNIST adversarial attack with more training samples.]{Accuracy and average entropy of CDNs and a BNN (Noisy K-FAC) under FGSM attack, with varying number of samples of $\vtheta$ used during training. Circles indicate accuracy, while crosses indicate entropy. The y-axis represents both the accuracy and the entropy relative to the maximum entropy (i.e.~$\ln 10$).
    While using more samples during training significantly improves  the overall performance of CDNs, it only has litle impact for BNNs (Noisy K-FAC and VMG).}
    \label{fig:adv_examples_more_samples}
\end{figure}

\begin{figure}[htb]
    \centering
    \subfloat[ML-CDN]{\includegraphics[width=0.5\linewidth]{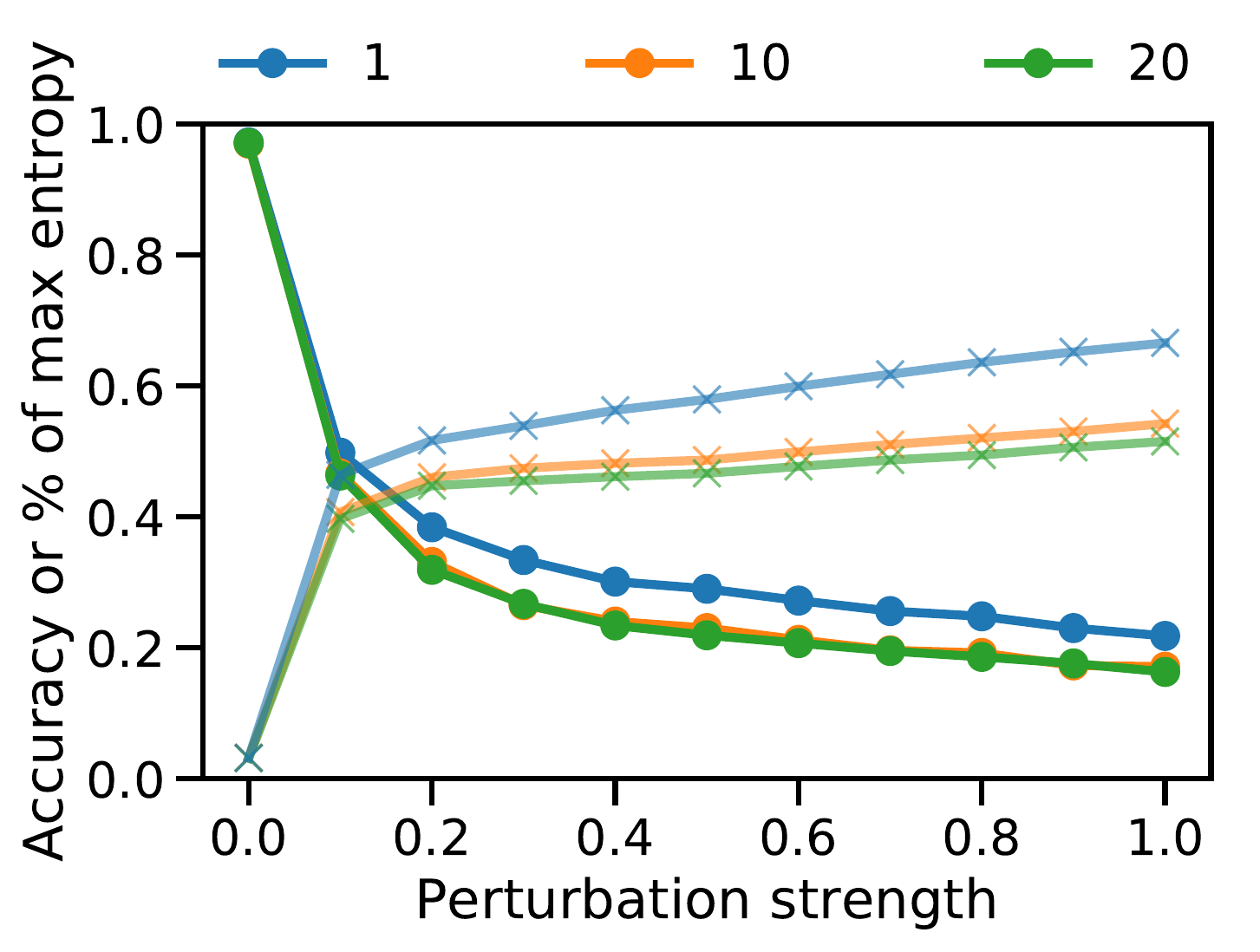}}
    \subfloat[VB-CDN]{\includegraphics[width=0.5\linewidth]{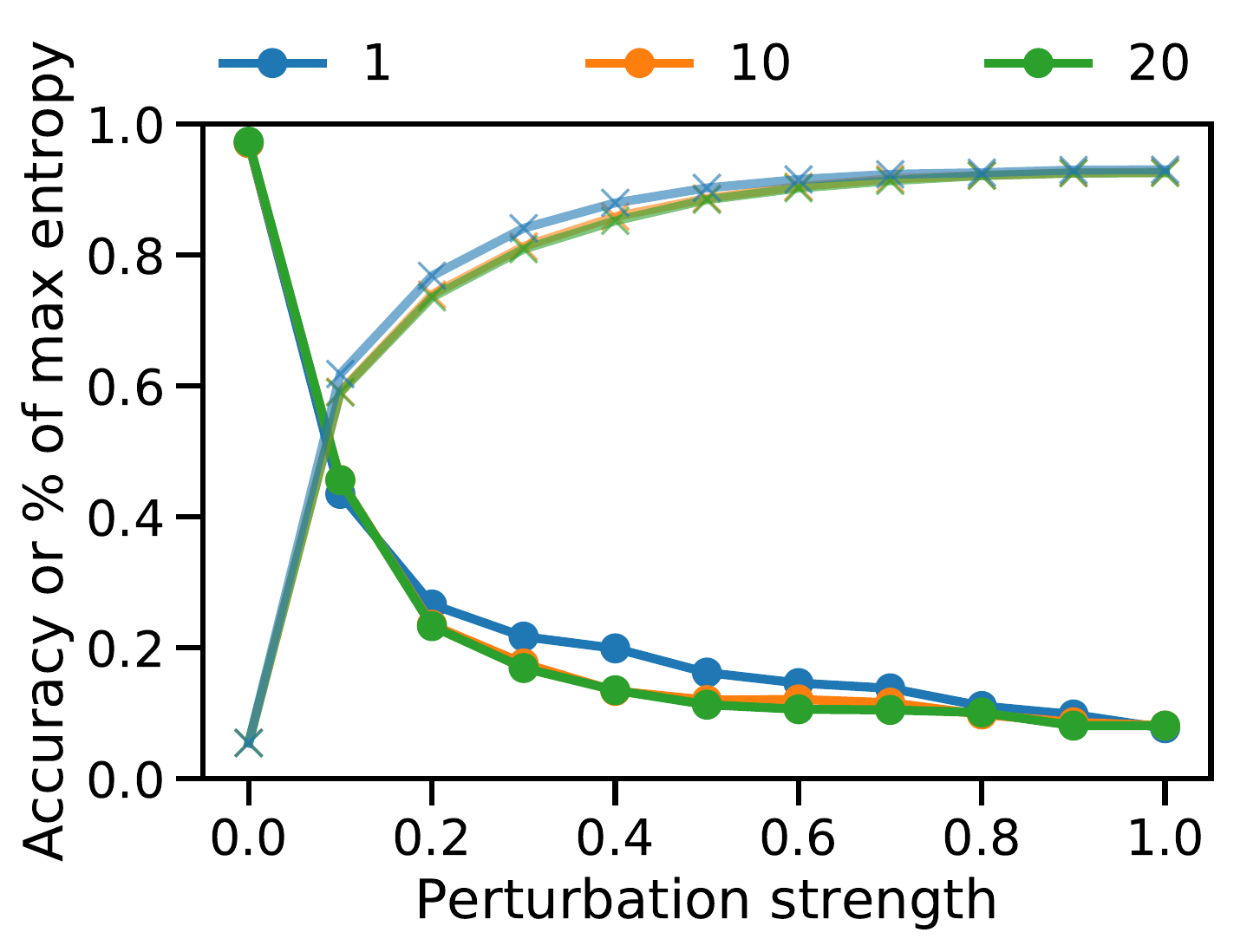}}
    
    \caption{Prediction accuracy and average entropy of CDNs for stronger adversarial examples, constructed by averaging over multiple forward-backward passes.}
    \label{fig:adv_examples_strong}
\end{figure}

\begin{figure}[htb]
    \centering
    \subfloat[Accuracy]{\includegraphics[width=0.5\linewidth]{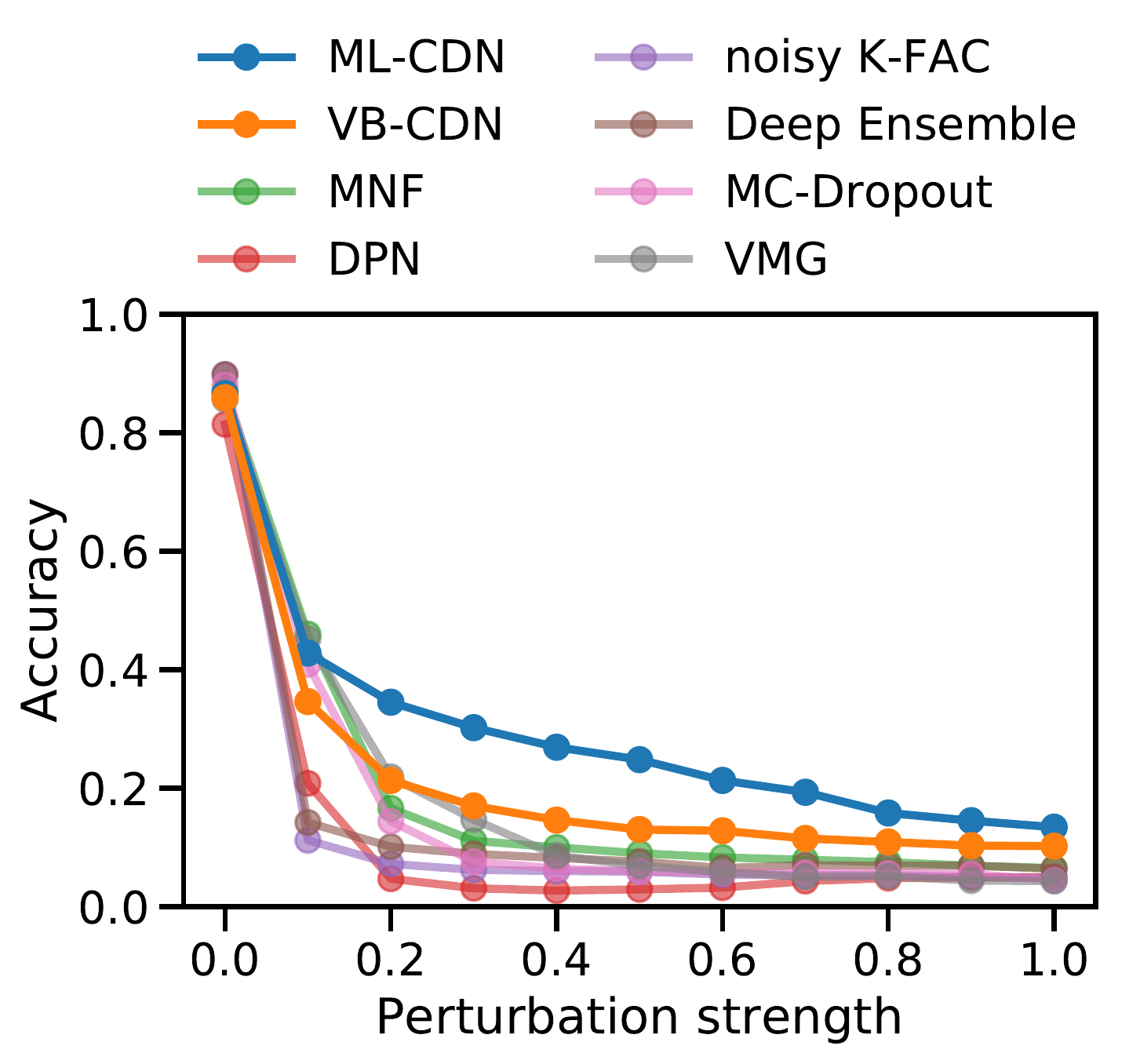}}
    \subfloat[Entropy]{\includegraphics[width=0.5\linewidth]{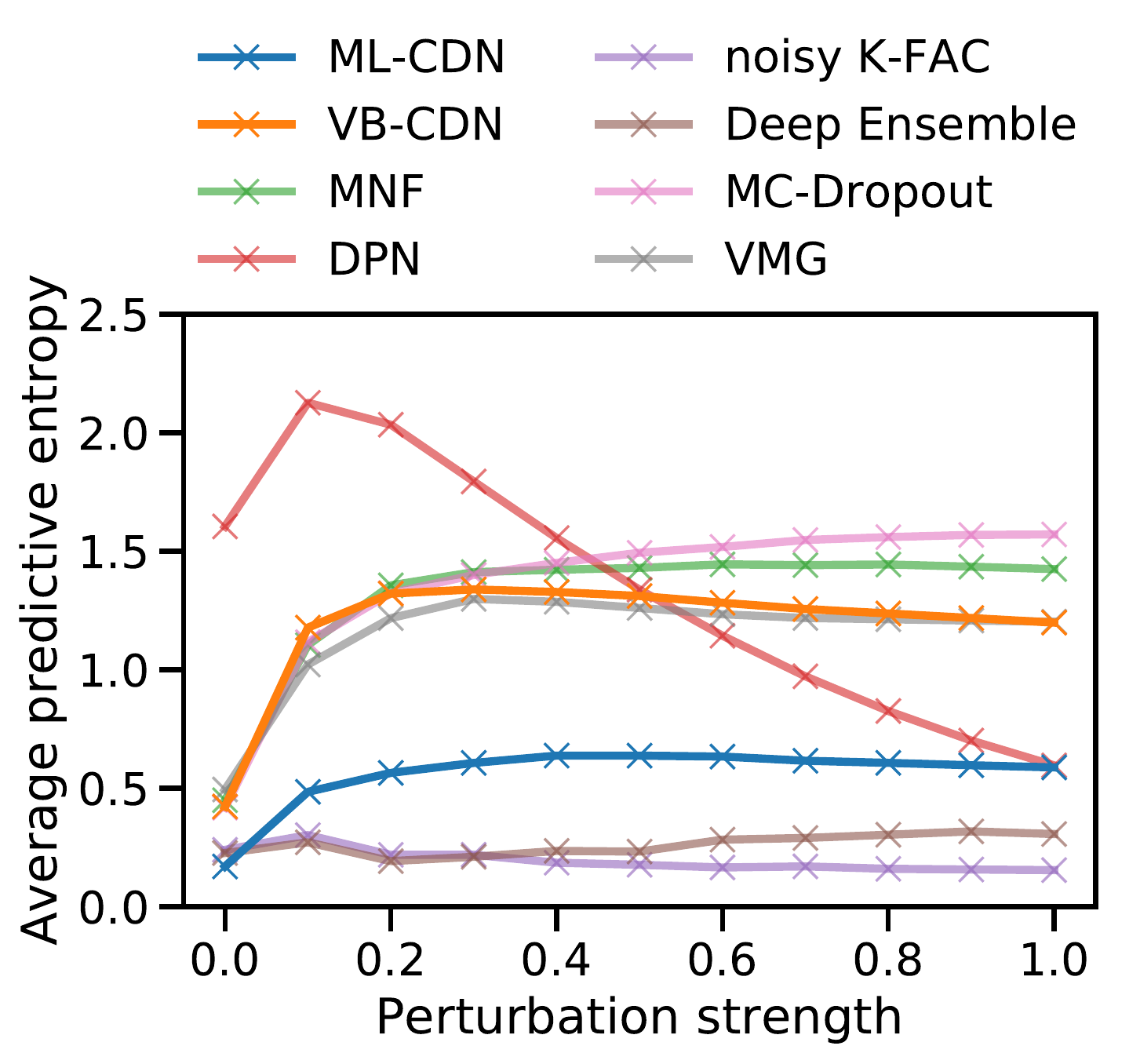}}
    
    \caption{Prediction accuracy and average entropy of models trained on Fashion-MNIST when attacked by FGSM-based adversarial examples with varying perturbation strength.}
    \label{fig:adv_examples_fmnist}
\end{figure}

    

\begin{figure}[htb]
    \centering
    \subfloat[Accuracy]{\includegraphics[width=0.5\linewidth]{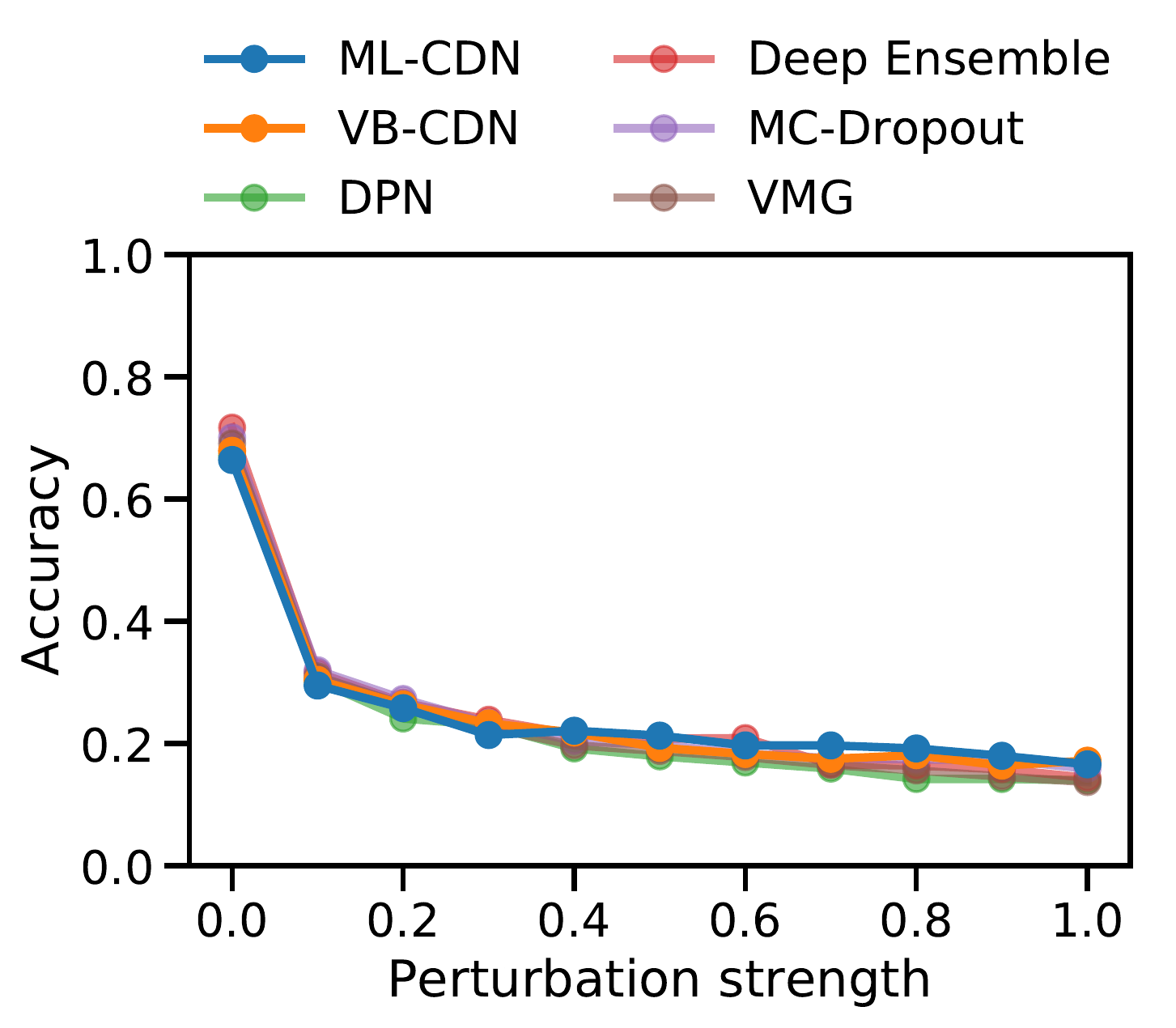}}
    \subfloat[Entropy]{\includegraphics[width=0.5\linewidth]{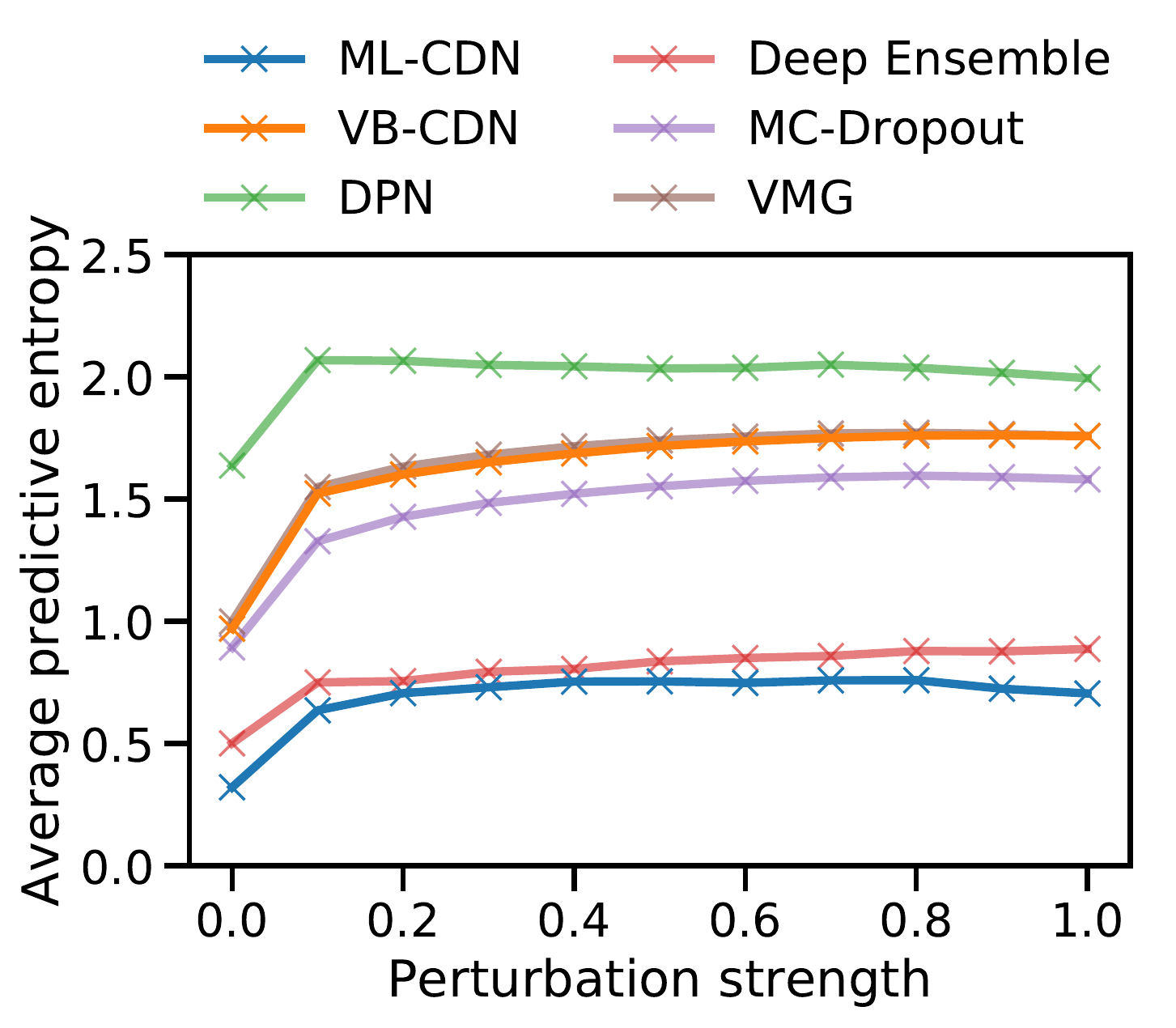}}
    
    \caption[CIFA-10 adversarial attack experiments.]{Prediction accuracy and average entropy of models trained on CIFAR-10 when attacked by FGSM-based adversarial examples with varying perturbation strength.}
    \label{fig:adv_examples_cifar}
\end{figure}

\subsection{Adversarial examples}
\label{subsec:adv_ex}

To investigate the robustness and detection performance of CDNs w.r.t.~adversarial examples~\citep{szegedy2013intriguing}, we apply the Fast Gradient Sign Method (FGSM)~\citep{goodfellow6572explaining} to a 10\% fraction (i.e.~1000 samples) of the MNIST, Fashion-MNIST \citep{xiao2017/online}, and CIFAR-10 test set.\footnote{We generate the adversarial examples based on a single forward-backward pass. 
} 
We do so, by making use of the  implementation provided by Cleverhans~\citep{papernot2018cleverhans}. 
We employ a transfer learning scheme by using DenseNet-121 \citep{huang2017densely} trained on ImageNet, as a fixed feature extractor for CIFAR-10.
The predictive network for both MNIST and CIFAR-10 is a two-layer MLP with 100 hidden units.
The probabilistic hypernetworks are two-layer MLPs with 50 hidden units. 
Note, that we do not use adversarial training when training the Deep Ensemble in this experiment to allow for a fair comparison. Furthermore, we use SVHN \citep{netzer2011reading} as the OOD training set for DPN baseline.

\textbf{MNIST:} \Cref{fig:adv_examples} presents the accuracy and the average empirical entropy of the predictive distribution w.r.t.~adversarial examples for MNIST with varying levels of perturbation strength (between 0 and 1). We observe that the CDNs are more robust to adversarial examples than all baseline models.
More specifically, the ML-CDN is significantly more robust in terms of accuracy to adversarial examples than all other models, while showing a competitive and nicely increasing entropy.
The VB-CDN has only slightly better prediction accuracy but attains higher uncertainty than all the baselines except the DPN.
Moreover, it shows uncertainties close to that of the DPN, while having higher accuracy and without 
needing additional data during training. 
Furthermore, we found that using more samples of $\vtheta$ during training is beneficial for the robustness of both ML-CDNs and VB-CDNs, as shown in \Cref{fig:adv_examples_more_samples}. This behavior is significantly more pronounced for CDNs than for BNNs (as exemplary shown for Noisy K-FAC and VMG). 
When using 10 samples per iteration during training the accuracy stays over $0.7$ and $0.5$ for ML-CDNs and VB-CDNs respectively, even for strong perturbations.
As shown in \Cref{fig:adv_examples_strong}, 
even when the adversarial examples are stronger, i.e.~estimated by averaging over multiple forward-backward passes, the performance of both CDNs is only marginally decreased (for VB-CDNs, it stays almost unchanged).



\textbf{Fashion-MNIST:}
The results on Fashion-MNIST are shown in \Cref{fig:adv_examples_fmnist}: Overall the same observations and conclusions can be made 
as for MNIST.
We note that strangely, the DPN's uncertainty estimate is decreasing with increasing perturbation strength.




\textbf{CIFAR-10:}
The results shown in \Cref{fig:adv_examples_cifar} demonstrate that the VB-CDN is competitive to other state-of-the-art models on CIFAR-10. The ML-CDN does not reflect uncertainty very well but has slightly higher accuracy than other models. 

\section{Conclusion}
\label{sec:conclusion}

We introduce compound density networks (CDNs), a new class of models that allows for better uncertainty quantification in neural networks (NNs) and corresponds to a compound distribution (i.e.~a mixture with uncountable components) in which both the component distribution and the input-dependent mixing distribution are parametrized by NNs. CDNs are inspired by the success of recently proposed ensemble methods and represent a continuous generalization of mixture density networks (MDNs) \citep{bishop1994mixture}. They can be implemented by using hypernetworks to map the input to a distribution over the parameters of the target NN, that models a predictive distribution. For training CDNs, regularized maximum likelihood or variational Bayes can be employed. Extensive experimental analyses showed that CDNs are able to produce promising results in terms of uncertainty quantification. Specifically, Bayesian CDNs are able to capture epistemic as well as aleatoric uncertainty, and yield very high uncertainty on out-of-distribution samples while still making high confidence predictions on within-distribution samples. Furthermore, when facing FGSM-based adversarial attacks, the predictions of CDNs are significantly more robust in terms of accuracy than those of previous models.
This robustness under adversarial attacks is especially pronounced for CDNs trained with a maximum likelihood objective, but also clearly visible for Bayesian CDNs, which also provide a better chance of detecting the attack by showing increased uncertainty compared to the baselines. 
These promising experimental results indicate the benefits of applying a mixture model approach in conjunction with Bayesian inference for uncertainty  quantification in NNs. We will investiage other implementations of CDNs and adaptions to recurrent and convolutional NNs in future.


\section*{Acknowledgements}
SD and AF are funded by the Deutsche Forschungsgemeinschaft (DFG, German Research Foundation) under Germany's Excellence Strategy - EXC 2092 CASA - 390781972.




\bibliography{icml2019_cdn}
\bibliographystyle{icml2019}

\end{document}